\documentclass[10pt,journal,compsoc]{IEEEtran}

\ifCLASSOPTIONcompsoc
  \usepackage[nocompress]{cite}
\else
  \usepackage{cite}
\fi

\usepackage{graphicx}
    \usepackage{booktabs}
\usepackage{tikz}
\usepackage{lipsum}
\usepackage{xcolor, soul}
\sethlcolor{yellow}
\usepackage{amsfonts,amssymb}
\usepackage{multirow}
\usepackage{enumitem}
\usepackage{amsmath}
\usepackage{comment}

\usepackage{hyperref}
\usepackage{pdfpages}

\begin{document}
\title{VisKnow: Constructing Visual Knowledge Base for Object Understanding}

\author{Ziwei~Yao,~\IEEEmembership{Student~Member,~IEEE,}
        Qiyang~Wan,~\IEEEmembership{Student~Member,~IEEE,}
        Ruiping~Wang*,~\IEEEmembership{Senior~Member,~IEEE,}
        and~Xilin~Chen,~\IEEEmembership{Fellow,~IEEE}% <-this % stops a space
\thanks{The authors are with the Key Laboratory of AI Safety of CAS, Institute of Computing Technology, Chinese Academy of Sciences (CAS),
Beijing 100190, China. E-mail: \{ziwei.yao, qiyang.wan\}@vipl.ict.ac.cn, \{wangruiping, xlchen\}@ict.ac.cn\protect\newline *~Corresponding author.}
}

\IEEEtitleabstractindextext{%
\begin{abstract}

Understanding objects is fundamental to computer vision.
Beyond object recognition that provides only a category label as typical output, in-depth object understanding represents a comprehensive perception of an object category, involving its components, appearance characteristics, inter-category relationships, contextual background knowledge, etc.
Developing such capability requires sufficient multi-modal data, including visual annotations such as parts, attributes, and co-occurrences for specific tasks, as well as textual knowledge to support high-level tasks like reasoning and question answering.
However, these data are generally task-oriented and not systematically organized enough to achieve the expected understanding of object categories.
In response, we propose the \textbf{Vis}ual \textbf{Know}ledge Base that structures multi-modal object knowledge as graphs, and present a construction framework named \textbf{VisKnow} that extracts multi-modal, object-level knowledge for object understanding.
This framework integrates enriched aligned text and image-source knowledge with region annotations at both object and part levels through a combination of expert design and large-scale model application.
As a specific case study, 
we construct \textbf{AnimalKB}, a structured animal knowledge base covering 406 animal categories, which contains 22K textual knowledge triplets extracted from encyclopedic documents, 420K images, and corresponding region annotations.
A series of experiments showcase how AnimalKB enhances object-level visual tasks such as zero-shot recognition and fine-grained VQA, and serves as challenging benchmarks for knowledge graph completion and part segmentation.
Our findings highlight the potential of automatically constructing visual knowledge bases to advance visual understanding and its practical applications.
The project page is available at \textcolor{blue}{\href{https://vipl-vsu.github.io/VisKnow}{https://vipl-vsu.github.io/VisKnow}}.
\end{abstract}

\begin{IEEEkeywords}
Knowledge Base Construction, Object Understanding, Visual Knowledge, LLM-assisted Application
\end{IEEEkeywords}}

\maketitle

\IEEEdisplaynontitleabstractindextext

\IEEEpeerreviewmaketitle

\ifCLASSOPTIONcompsoc
\IEEEraisesectionheading{\section{Introduction}\label{sec:introduction}}
\else
\section{Introduction}
\label{sec:introduction}
\fi

\IEEEPARstart {U}{nderstanding} objects is a fundamental requirement in computer vision systems~\cite{palmeri2004visual}.
The field has witnessed significant breakthroughs in the area of object recognition, enabling AI models to identify and classify a wide array of everyday items with increasing accuracy.
However, a truly in-depth understanding of object categories extends beyond classification to encompass semantic aspects such as constituent components, appearance characteristics, inter-category relationships, and relevant contextual background.
These factors, which either define the category itself or describe how it interacts with other categories and the environment, necessitate comprehensive multi-modal knowledge.
Incorporating such explicit knowledge is highly beneficial even for pre-trained large language models, as it significantly mitigates hallucinations, enhances reasoning capability, and improves interpretability, especially in specialized domains or scenarios demanding high robustness~\cite{zhang2019ernie,pan2024unifying}.

\begin{figure}[t]
    \centering
    \includegraphics[width=0.95\linewidth]{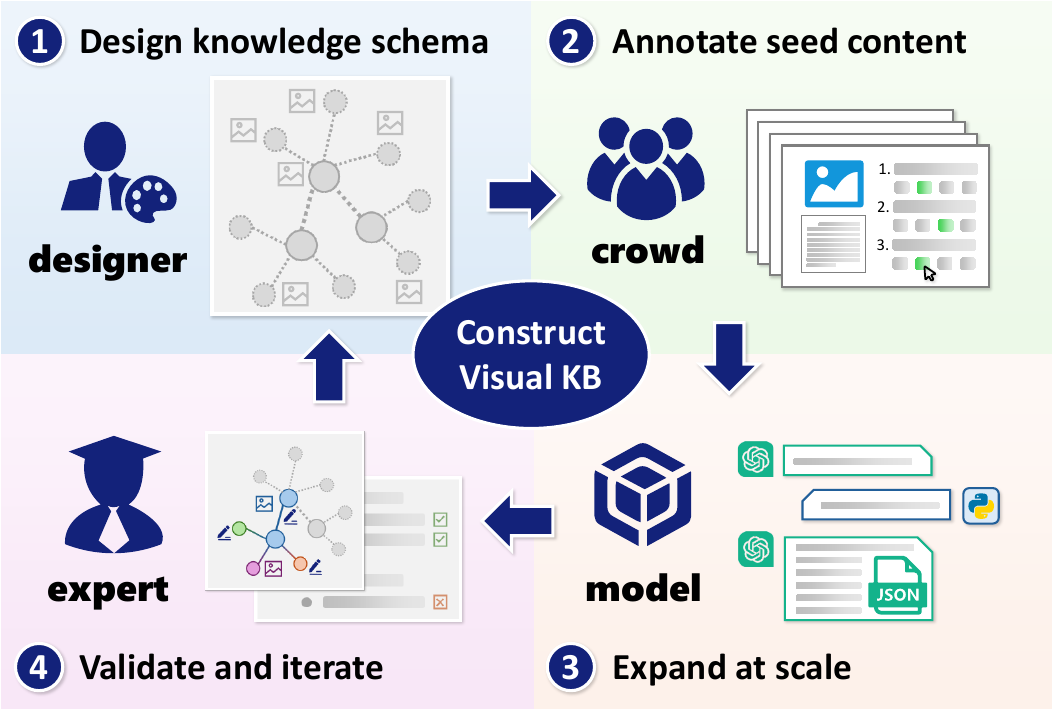}
    \caption{The proposed visual knowledge base is developed through a series of cascading steps: 1) designing knowledge schema, 2) annotating seed content, 3) expanding at scale, and 4) validating and iterating. To minimize construction costs, expert involvement is restricted to essential design and validation tasks, while data collection is delegated to crowdsourcing or automated models.}
    \label{fig:real_title}
\end{figure}

\begin{figure*}[t]
\begin{center}
  \includegraphics[width=0.92\linewidth]{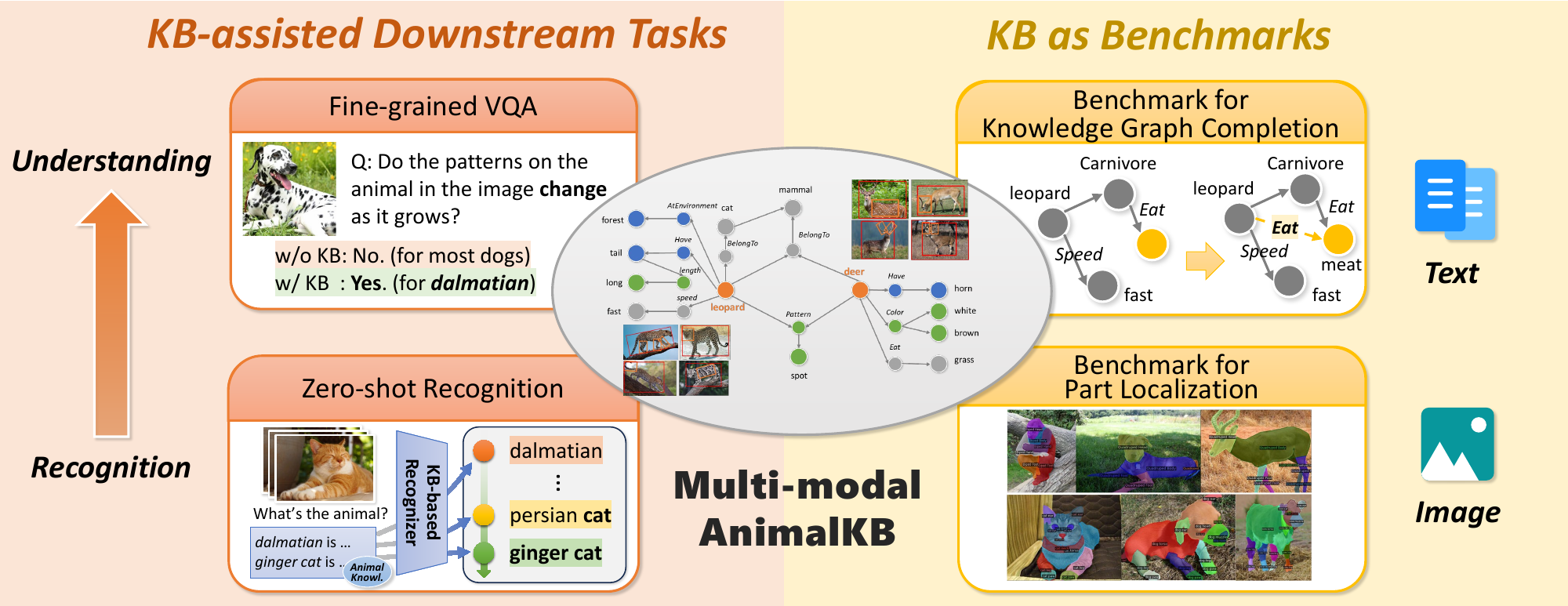}
\end{center}
  \caption{Multi-modal \textbf{AnimalKB} constructed by the proposed \textbf{VisKnow} framework can be applied in various aspects, including enhancing knowledge-related visual tasks, and providing annotations required for constructing benchmarks.}
\label{fig:title}
\end{figure*}

Explicit knowledge refers to structured symbolic descriptions of specific objects, typically presented in formats such as labels, text, or triplets~\cite{ji2021survey,sheth2019shades}.
\textbf{Visual knowledge} refers to a part of knowledge that directly describes visual content, which is essential for a deep understanding of objects, including their attributes, parts, affordances, co-occurrences, and appearance similarities.
Common recognition datasets generally contain the most basic visual knowledge, such as category labels in ImageNet~\cite{deng2009imagenet}, providing minimal visual information for classification tasks.
There have been some preliminary attempts to develop deeper visual knowledge for specific domains, such as birds~\cite{WahCUB_200_2011} and other animals~\cite{lampert2009learning}.
These domain-specific datasets contain semantic labels of cross-category visual concepts such as attributes and parts.
These multi-granularity annotations are designed for certain tasks to enhance object-level understanding in those specific areas.
On the other hand, generally well-known knowledge bases mainly comprise \textbf{non-visual knowledge} extracted from visual-agnostic corpora, such as WordNet~\cite{miller1995wordnet}, ConceptNet~\cite{liu2004conceptnet}, and ATOMIC~\cite{sap2019atomic}.
They focus on the summary and explanation of concepts, common knowledge, and behaviors within human society,
encompassing the information that is not directly related to the visual content.
They are frequently employed to support high-level tasks, such as reasoning and question answering, where it is typically assumed that basic object recognition has already been effectively achieved.
As a result, we observe that previous visual knowledge has often been object-centric but task-oriented, concentrating on a specific format or particular task; in contrast, non-visual knowledge tends to be more general and abstract, focusing on broader viewpoints but overlook object-level details.
To achieve in-depth object understanding from visual appearance to high-level semantics, it is therefore crucial to jointly exploit and align both visual and non-visual knowledge in a consistent way.

This paper proposes a systematic, object-centric knowledge structure that effectively integrates the multi-modal knowledge, facilitating its application in downstream vision tasks related to object understanding.
In addition, we propose \textbf{VisKnow}, a streamlined framework for constructing such comprehensive object-level visual knowledge base.
As shown in Fig.~\ref{fig:real_title}, the framework utilizes standardized and sequential processes to effectively integrate crowdsourcing and large-scale pre-trained models into the construction of the visual knowledge base, thereby minimizing expert involvement and reducing associated costs.
Experts are solely tasked with developing essential knowledge structures and construction rules, along with conducting limited validation to ensure quality.
Technically, based on the collected text- and image-source data, experts initially define the knowledge schema and specific categories of interest.
We then develop a cost-effective crowdsourcing annotation system to gather seed data.
Next, large language models and visual grounding models~\cite{cheng2024yolo,kirillov2023segment} are employed to expand the knowledge scale and achieve multi-modal alignment.
Finally, experts validate these annotations, making iterative adjustments to ensure quality.
As shown in Fig.~\ref{fig:title}, the constructed visual knowledge base provides the multi-granularity knowledge required for various visual tasks, which includes the following characteristics:
(1) specific and informative \textbf{text-based} knowledge, (2) object- and part-level visual concepts with \textbf{region annotations}, and (3) robust \textbf{alignment} between the visual and textual modality.

As a specific case study, we develop an object-level visual knowledge base focusing on animal categories by leveraging the proposed VisKnow.
We collect encyclopedic articles for 406 animal species and over 400K corresponding images from the Internet and existing visual datasets.
Consequently, a multi-modal, multi-grained animal knowledge base, referred to as \textbf{AnimalKB}, is constructed, which comprises more than 22K knowledge entries and 480K region annotations.
We demonstrate the utility of AnimalKB through a series of experiments, highlighting its application in downstream tasks and its effectiveness as benchmarks.

\begin{itemize}[leftmargin=*]
\item \textbf{KB-assisted downstream tasks} (Sec.~\ref{sec:app}) highlights the contribution of additional knowledge from AnimalKB to enhance visual model capabilities:
\end{itemize}
\begin{enumerate}[leftmargin=*]
\item \textbf{Zero-shot Recognition}: AnimalKB is used to improve the recognition of fine-grained categories in image-text matching models, such as CLIP~\cite{radford2021learning}.
\item \textbf{Fine-grained VQA}: AnimalKB bolsters the ability of MLLMs to answer professional questions about fine-grained categories.
\end{enumerate}

\begin{itemize}[leftmargin=*]
\item \textbf{KB as benchmarks} (Sec.~\ref{sec:bench}) demonstrates how the comprehensive annotations of AnimalKB can be used to evaluate existing specialized tasks:
\end{itemize}
\begin{enumerate}[leftmargin=*]
\item \textbf{TextBench}: Text annotations from AnimalKB are used to assess the effectiveness of knowledge graph completion models in processing encyclopedic knowledge.
\item \textbf{PartBench}: Part annotations from AnimalKB are utilized to evaluate part segmentation models across a broader range of animal categories.
\end{enumerate}

In addition, a KB-guided interpretation system is designed to illustrate the use of AnimalKB in practical scenario such as museums and nature reserves. 
The experiments and use cases demonstrate the substantial potential of the proposed multi-modal knowledge construction framework for enhanced visual understanding.

\section{Related Work}

\begin{figure}[t]
\begin{center}
  \includegraphics[width=0.9\linewidth]{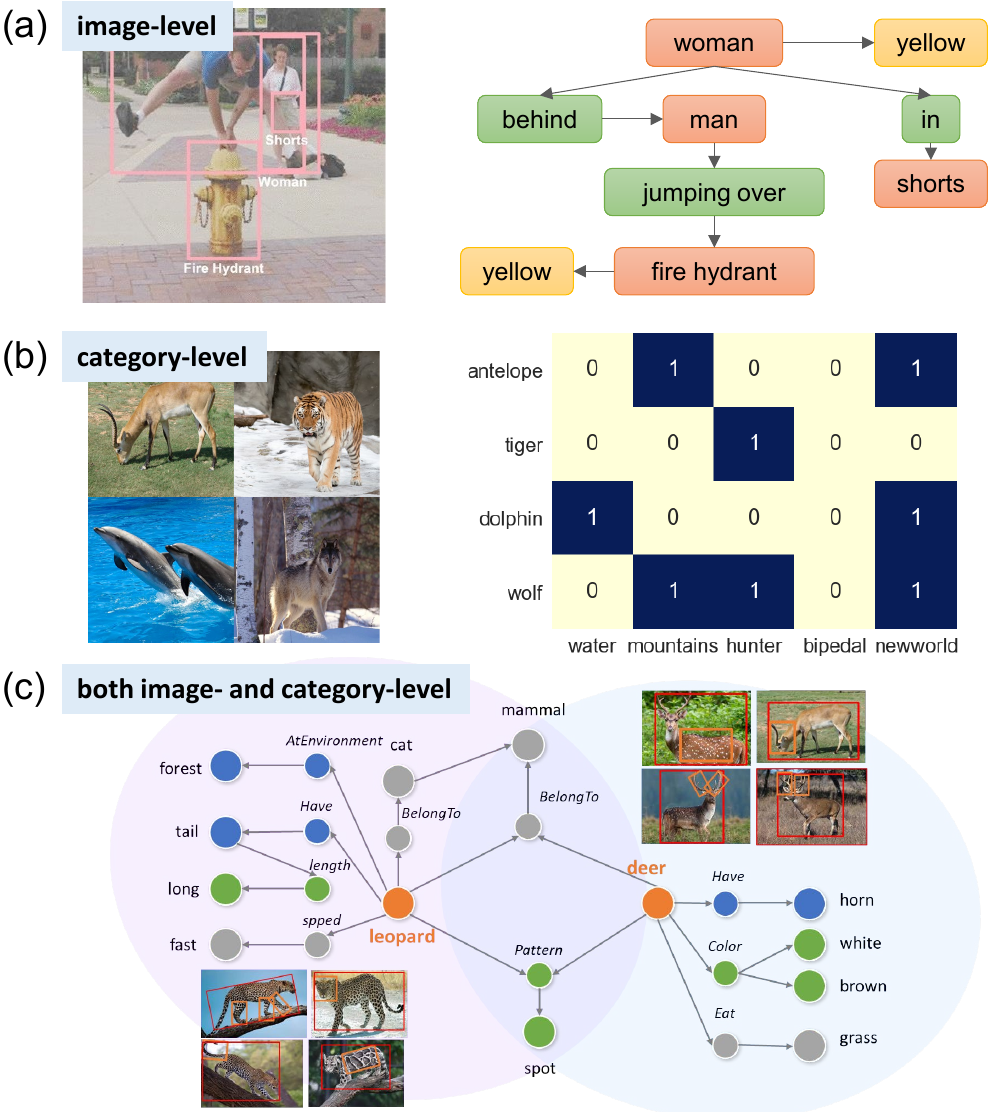}
\end{center}
  \caption{Illustration of various types of visual datasets: (a) image-level annotations, such as bounding boxes and scene graphs in Visual Genome~\cite{krishna2017visual}; (b) category-level annotations, such as attribute annotations in AwA2~\cite{lampert2009learning}; and (c) datasets that provide both image- and category-level annotations, such as the proposed visual knowledge base with its rich, multi-modal semantic information.}
\label{fig:visual_dataset}
\end{figure}

\subsection{Knowledge in Visual Datasets}

The construction of diversified visual datasets has driven the development of the field of computer vision. The association between visual images and category labels in these datasets contains the most fundamental visual knowledge.
The introduction of ImageNet~\cite{deng2009imagenet} marks a transformative milestone in the field by significantly expanding the range of object recognition categories to a large scale.
Such datasets~\cite{kuznetsova2020open,VanHorn2018iNaturalist,zhou2017places} are fully annotated with category labels, thereby facilitating supervised learning for models engaged in recognition tasks.

In order to enhance visual understanding beyond recognition, more tasks have been proposed, such as detection~\cite{lin2014microsoft,everingham2010pascal}, segmentation~\cite{lin2014microsoft,cordts2016cityscapes}, and visual relationship detection~\cite{krishna2017visual,lu2016visual}.
The datasets contain detailed annotations for image samples beyond simple category labels, such as bounding boxes, segmentation masks, attributes, and visual relationships (Fig.~\ref{fig:visual_dataset}(a)).
These annotations facilitate advancements in scene understanding, exemplified by techniques such as scene graph generation~\cite{lu2016visual,yao2018scene,zellers2018neural}, which represents scenes as graph structures.
In recent years, large-scale datasets consisting of image-text pairs~\cite{schuhmann2021laion,changpinyo2021conceptual,gadre2023datacomp} have significantly advanced the development of multi-modal pre-training models, 
enabling the creation of models like CLIP~\cite{radford2021learning} that perform exceptionally well in multi-modal alignment.
These models have established a vital foundation for vision-language models~\cite{li2023blip2,li2023llava}.

As visual datasets grow in diversity and scale, visual models built on these datasets are becoming increasingly powerful, enhancing their ability to generalize across common objects and scenes.
However, specialized tasks, such as fine-grained recognition or interpretable robust recognition, necessitate more diverse and customized category-level knowledge in the specific domain.
Current pre-trained visual models often perform suboptimally on such tasks due to the limited visual knowledge beyond category labels.
These application scenarios typically demand strong specialization and face limited data availability, necessitating visual models that incorporate expert knowledge, such as visual attributes~\cite{lampert2009learning,WahCUB_200_2011}
on the target categories (Fig.~\ref{fig:visual_dataset}(b)).
This paper explores methods for constructing and applying detailed, multi-level visual knowledge bases to address these challenges (Fig.~\ref{fig:visual_dataset}(c)).

\begin{figure*}[t]
      \centering
      \includegraphics[width=2.0\columnwidth]{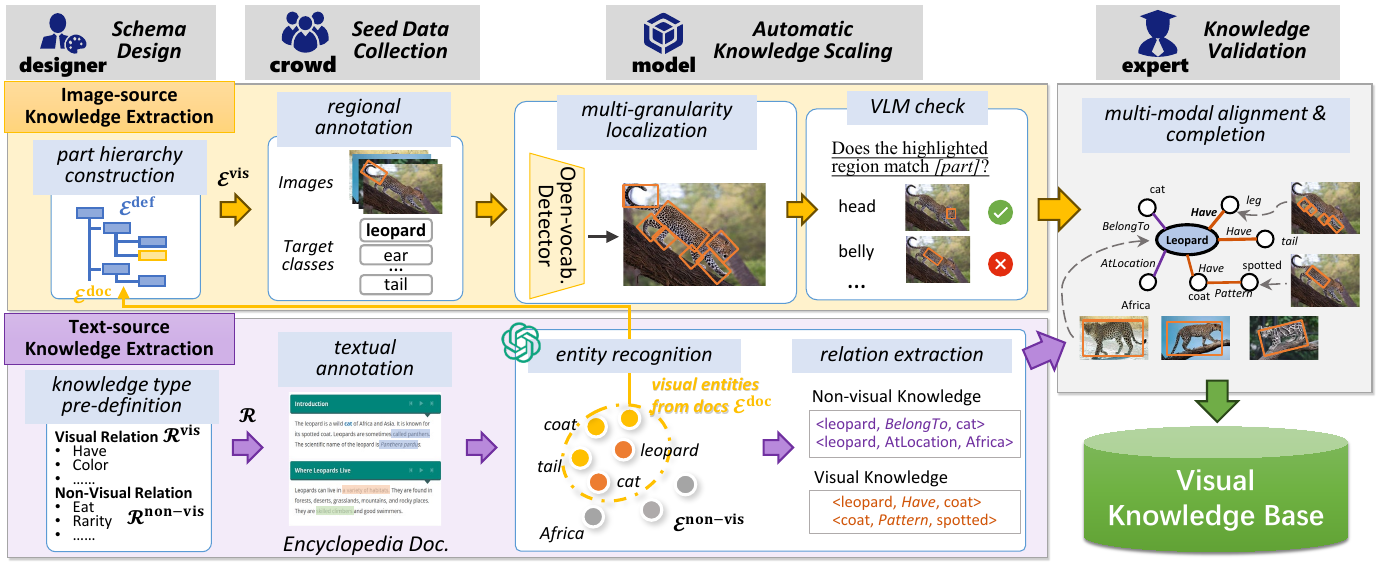}
      \caption{The framework of the proposed VisKnow consists of four stages, in which text-source and image-source knowledge are extracted separately. Given encyclopedia documents, the LLM extracts entities and relations from the text in a structured format under the guidance of pre-defined knowledge types and a few manual annotations. Visual entities are then fed back into the knowledge schema design to refine the part hierarchy, and regional annotations are applied to the seed images accordingly. For image-source knowledge extraction, we utilize and finetune detectors for object and part localization and verify its results with the Vision-Language Models. Finally, the image regions are aligned with text-source knowledge nodes and incorporated into the visual knowledge base.}
      \label{fig:pipieline}
\end{figure*}

\subsection{Multi-modal Knowledge Graph}

Knowledge graphs representing knowledge as graphs to illustrate associations between concepts, have long played a vital role in the AI area. Typical works on knowledge graph include WordNet~\cite{miller1995wordnet}, DBpedia~\cite{auer2007dbpedia}, Wikidata~\cite{vrandevcic2014wikidata}, ConceptNet~\cite{speer2017conceptnet}, and WebChild~\cite{tandon2017webchild}, supporting many applications such as reasoning, question answering, and recommendation systems.
Recognizing the crucial role of visual representation in human cognition, researchers have started to focus on how to integrate multi-modal information into knowledge graphs. 
According to~\cite{Zhu2022multimodalKG,Chen2024KGMM}, existing multi-modal knowledge graphs (MMKGs) can be categorized into A-MMKG, which treats images as attributes of entities, and N-MMKG, which treats images as independent entity nodes associated with other entities through relationships like \textit{imageOf}.
As N-MMKGs, IMGpedia~\cite{ferres2017imgpedia} connects visual data from Wikimedia Commons with metadata from DBpedia\cite{auer2007dbpedia}; VisualSem\cite{visualsem2021} filters entities and images from BabelNet\cite{babelnet2012}, ensuring high quality and diversity; Richpedia~\cite{zhang2020richpedia} gathers images and descriptions from Wikipedia and offers hyperlinks for relationship identification.
NEIL~\cite{chen2013neil}, and GAIA~\cite{gao2018gaia} build MMKGs through labeling images, also belonging to the N-MMKG.
As A-MMKGs, ImageGraph~\cite{onoro2017answering} and MMKG\cite{liu2019mmkg} collect image entities using web crawlers based on Freebase\cite{Bollacker2008Freebase}. 

Though these MMKGs attach visual representations to textual concepts, most of them focus on world knowledge and pay limited attention to object- and part-level knowledge, thus offering restricted assistance for visual tasks.
The proposed VisKnow aims to establish a visual knowledge base that systematically integrates textual knowledge with images and fine-grained region annotations, providing substantial support for object understanding.

\subsection{Automatic Knowledge Base Construction}

The construction of knowledge graphs is moving from previous manual building towards a highly automated direction. 
Early knowledge graphs like Cyc~\cite{lenat1989building} and WordNet~\cite{miller1992wordnet} relying on experts are accurate but costly and limited in scalability.
Then, projects like Freebase~\cite{Bollacker2008Freebase}, ConceptNet~\cite{liu2004conceptnet} and Wikidata~\cite{vrandevcic2014wikidata} utilized crowdsourcing for scalable, multi-domain knowledge collaboration.
Marking an early attempt at automatic knowledge acquisition, NELL system~\cite{carlson2010toward} continuously and dynamically learns entities and relations from the web, and
NEIL~\cite{chen2013neil} is considered as its multi-modal evolution.

With the development of deep learning, neural network techniques such as knowledge embedding~\cite{bordes2013translating,trouillon2016complex,nickel2016simple} and graph neural networks~\cite{yao2019graph} significantly enhance the construction and completion of knowledge graphs. 
Moreover, pre-trained language models (e.g. BERT~\cite{devlin2019bert}, GPTs~\cite{radford2019gpt2}, etc.) have demonstrated strong semantic understanding capabilities in knowledge extraction tasks such as entity recognition and relation extraction~\cite{lin2020bert,eberts2019spert}. 
COMET~\cite{bosselut2019comet} pre-trains Transformer models on large-scale text data, and then uses entities and specific relations as ``probes'' to extract relevant knowledge entries in the model.
In MMKG construction, GAIA~\cite{gao2018gaia} integrates text and visual knowledge extraction for complex graph queries and multimedia retrieval, while UKnow~\cite{uknow2020} uses a unified protocol with pre-trained models to categorize and aggregate data into a graph.
These methods rely on manually annotated data for training knowledge extraction models, and the knowledge classes are usually fixed, making them difficult to expand during construction.
Advanced LLMs lower the barrier to constructing knowledge graphs by using prompts to directly extract triplets from text with minimal examples.
AutoKG~\cite{chen2023autokg} leverages an LLM to automatically extract key concepts and drive lightweight knowledge-graph construction.graph.
$\text{M}^\text{2}\text{ConceptBase}$~\cite{m2conceptbase2023} extracts and aligns concepts from image-text pairs, and uses the LLM for generating supplementary descriptions.
Directly mining entities and relations from raw data reduces human cost, but the quality of the constructed knowledge graph is inevitably affected by noise and ambiguity in the original data. 

Therefore, our proposed VisKnow introduces a cascading framework, in which experts, crowdsourced workers, and models collaborate to reduce manual annotation costs while gradually expanding the knowledge base with controllable quality.
The human-in-the-loop approach makes it easier to find the balance between manual quality assurance and model-driven scalability in multiple iterations.
In addition, with regard to visual knowledge acquisition, compared to other works that only collect images or coarse object-level regions, we leverage detectors and segmentors to obtain fine-grained part-level region annotations.

\section{Construction Approach}
This section introduces the proposed \textbf{VisKnow} framework for constructing knowledge bases.
Using animal categories as a representative case, we construct AnimalKB to demonstrate the framework's capabilities in organizing and managing multi-modal data.
We first introduce the formalization of the visual knowledge base in Sec.~\ref{approach_def} and the data collection process in Sec.~\ref{approach_data}. 
Next, as shown in Fig.~\ref{fig:pipieline}, we extract knowledge from both text and image sources. 
The construction pipeline comprises \textbf{expert schema design}, \textbf{crowdsourced seed data annotation}, \textbf{model-based knowledge scaling}, and \textbf{iterative verification and refinement}, which will be detailed in Sec.~\ref{approach_text} for textual modality, and Sec.~\ref{approach_vis} for visual modality. 
Finally, in Sec.~\ref{approach_align}, we align and complete the multi-modal knowledge to obtain a comprehensive knowledge base.

\subsection{Formalization of the Visual Knowledge Base}
\label{approach_def}
We formalize the visual knowledge base as a graph $\mathcal{G}$ represented by triplets:
\begin{equation}
\mathcal{G}=\left\{\left(h, r, t\right)|h,t\in\mathcal{E},r\in\mathcal{R}\right\},
\label{eq:triple}
\end{equation}
where each triplet represents a knowledge entry, composed of a head entity $h$ and a tail entity $t$ from the entity set $\mathcal{E}$, and $r \in \mathcal{R}$ referring to the relation between $h$ and $t$.

According to whether they can be directly represented and learned in visual modality, the entities $\mathcal{E}$ and relations $\mathcal{R}$ are categorized into separate \emph{visual} and \emph{non-visual} subsets for convenience.
Visual entities $\mathcal{E}^{\text{vis}}$ represent entities such as animals, visible parts, and appearance attributes, 
while non-visual entities $\mathcal{E}^{\text{non-vis}}$ encode abstract or unobservable concepts. 
Similarly, visual relations $\mathcal{R}^\text{vis}$ denote observable predicates
such as \texttt{Have}, \texttt{Color}, and \texttt{Num}, 
whereas non-visual relations $\mathcal{R}^\text{non-vis}$ describe abstract or conceptual properties 
such as \texttt{BelongTo}, \texttt{Synonym}, and \texttt{Eat}.

Specifically, a non-visual entity in $\mathcal{E}^\text{non-vis}$ is described by a single textual label $\ell_e$, whereas a visual entity $e^\text{vis} \in \mathcal{E}^\text{vis}$ contains more:
\begin{equation}
e^\text{vis} = \left( \ell_e, \{ (i_k, a_k) \}_{k=1}^m \right),
\label{eq:visual_node}
\end{equation}
where $\ell_e$ is the textual label of the entity, and 
$\{ (i_k, a_k) \}_{k=1}^m$ is a set of $m$ images with the corresponding area annotations.
Each $i_k$ is an image where the entity occurs, 
and each $a_k$ indicates the entity's location within the image, such as a bounding box or segmentation mask.

Similarly, a non-visual relation in $\mathcal{R}^\text{non-vis}$ is described by a single textual label $\ell_r$, and each visual relation $r^\text{vis} \in \mathcal{R}^\text{vis}$ is represented as:
\begin{equation}
r^\text{vis} = \left( \ell_r, \{ (i_k, a_k) \}_{k=1}^n \right),
\label{eq:relation}
\end{equation}
where $\ell_r$ is the textual label of the relation, 
and $\{ (i_k, a_k) \}_{k=1}^n$ is a set of $n$ image–region annotation pairs. 
Each $i_k$ is an image in which the relation is visually grounded, 
and $a_k$ is the union area of its head and tail entities in $i_k$.
Triplets with visual relations are termed as \textbf{visual knowledge}, and others are referred to as \textbf{non-visual knowledge}.

\subsection{Data Collection}
\label{approach_data}

The proposed visual knowledge base integrates multi-modal data, encompassing both textual and visual modalities, distinguishing it from standard knowledge bases that contain only textual data.

In textual modality, traditional data sources such as Wikipedia~\cite{wikipedia} are commonly used.
While these sources contain abundant knowledge, they lack a visual-centric focus essential for a visual knowledge base, which demands detailed and comprehensive visual descriptions that are typically absent in encyclopedias.
In contrast, educational corpora aimed at younger audiences often include the fundamental visual descriptions.
As an example, we select animal categories and utilize encyclopedia entries of 406 species from the Encyclopedia Britannica Kids~\cite{britannica_kids} as the textual knowledge source.
These selected categories cover species such as mammals, birds, reptiles and arthropods, offering much broader coverage than previous domain-specific datasets such as AwA2~\cite{Xian2018ZeroShotLearning} and CUB~\cite{WahCUB_200_2011}.

In visual modality, it is essential to collect a substantial number of high-quality images that align with the textual modality.
Initially, we acquire images from multiple visual datasets, including AwA2~\cite{lampert2009learning}, ImageNet~\cite{deng2009imagenet}, iNaturalist~\cite{VanHorn2018iNaturalist}, and OpenImages~\cite{kuznetsova2020open}.
For categories with insufficient samples, we augment the collection with additional images from the iNaturalist website and search engines.
To ensure both visual clarity and effective alignment with textual knowledge, we select images according to two criteria: (1) moderate resolution and clarity, and (2) clearly visible main subject animal.

It is important to note that here the mentioned data sources are selected specifically for our case study to build AnimalKB.
In other fields, data sources should be chosen based on specific requirements.
High-quality data sources can always simplify the data processing workflow and thus require careful selection by domain experts.

\subsection{Text-source Knowledge Extraction}
\label{approach_text}

Knowledge base construction is always domain-specific, rendering most general tools inapplicable without customization.
Consequently, expert-generated seed data is essential to maintain quality.
The proposed pipeline aims to reduce human experts' workload while leveraging a cascade construction flow including expert schema design, crowdsourced seed data annotation, and model-based knowledge scaling.
As illustrated by the purple block in Fig.~\ref{fig:pipieline}, we begin by extracting knowledge triplets from animal encyclopedia documents.
These triplets serve as the backbone of the knowledge base, referred to as $\mathcal{G}_{text}$.

Based on the categories and documents, experts first design the knowledge schema and detailed annotation guidelines, and initially define a set of knowledge types $\mathcal{R}$, which are further divided into visual relations $\mathcal{R}^{\text{vis}}$ and non-visual relations $\mathcal{R}^{\text{non-vis}}$.
For each document, the topic animal is designated as root node. All other nodes are suggested to be connected to it through one or more edges, thereby constructing a fully connected graph.

Then, to obtain seed knowledge annotation, human annotators are asked to extract knowledge triplets from the text.
We develop a textual knowledge annotation system, which performs semantic analysis of the original text based on rules, and constrain the consistency of entities and relations, allowing crowdsource to more easily complete the task. Annotation rules are refined and iterated based on their feedback.

Subsequently, the seed knowledge is scaled up using the LLM-based method, involving entity recognition and relation extraction.
Given an animal document $D$, we first divide it into multiple segments $\{d_1, d_2, \ldots, d_n\}$ based on paragraphs. 
For each segment, knowledge triplets are extracted independently to avoid information loss caused by overly long contextual inputs.
We concatenate the segment with few-shot examples to form a prompt, which is input into the LLM for generating knowledge triplets, involving entity recognition and relation extraction.
During entity recognition, entities that represent \textit{animals} and \textit{parts} are additionally incorporated into the set $\mathcal{E}^{\text{doc}}$, serving as a part of targets in the subsequent image-source knowledge extraction (see Sec.~\ref{approach_vis}).
For relation extraction, we provide pre-defined knowledge set $\mathcal{R}$ in the prompt, and ask the LLM to categorize each triplet as visual or non-visual. 
Experts validate the LLM’s output and iteratively adjust the pre-defined set for 2–3 rounds until the results stabilize.
Finally, for each document \( D \), the extracted triplets from all segments are aggregated into a fully-connected textual knowledge graph $\mathcal{G}_{text}$.
More details are provided in the supplementary material.

\subsection{Image-source Knowledge Extraction}
\label{approach_vis}

\begin{figure}[t]
    \centering
    \includegraphics[width=0.9\linewidth]{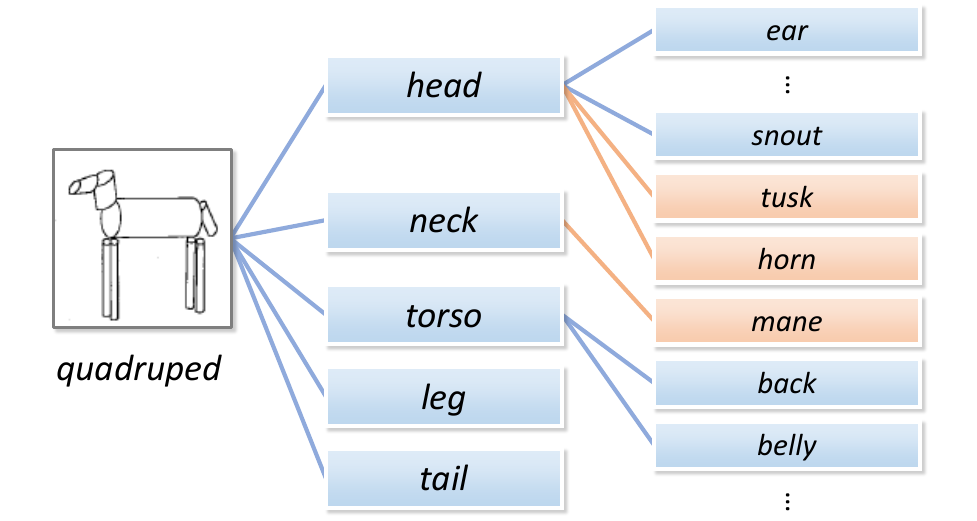}
    \caption{Experts design a general hierarchy for the parent category of animals (in blue), and then for each animal, supplement it with the key parts extracted from the textual knowledge (in orange), thus obtaining the part hierarchy.}
    \label{fig:part_structure}
\end{figure}

Enriching the visual entities and relations in the knowledge base with visual resources yields a more intuitive multi-modal knowledge system, which can further contribute to fine-grained and transferable visual object understanding.
As visual entities are typically associated with whole animals or their constituent parts, corresponding region-level annotations are additionally incorporated into these entities alongside animal images, as illustrated by the orange block in Fig.~\ref{fig:pipieline}.
Similar to text-source knowledge, image-source knowledge extraction undergoes a sequential cascade process, consisting of expert design, crowdsourced seed collection, and model scaling.

To obtain the region categories to be extracted, as illustrated in Fig.~\ref{fig:part_structure}, a general entity set $\mathcal{E}^\text{def}$ for each animal is pre-defined according to the part-level hierarchy of its supercategory, with the animal itself serving as the root.
Subsequently, the unique visual entities $\mathcal{E}^\text{doc}$ extracted from textual documents are inserted into $\mathcal{E}^\text{def}$, resulting in a complete and structured list of target entities $\mathcal{E}^\text{vis}$.

To collect seed region annotation, we first develop an annotation system to display, draw and adjust bounding boxes.
Specifically, we first group animal categories into several larger groups based on appearance similarities, select seed images for each group, and annotate the part regions with bounding boxes. Then, we train an object segmentation model~\cite{He_2017_ICCV} to perform inference on more images. After checking the test images, we incorporate them back into the training set and iterate on the segmentation model to improve its performance. The aforementioned process is conducted for multiple rounds. Finally, for images that are automatically annotated with parts, we manually review and modify them to ensure both the quantity and quality. The annotation and verification process is human-intensive, and thus difficult to apply to the large-scale scenario.

In order to further improve the efficiency of region annotation, we utilize the latest open-vocabulary object detection methods, such as GLIP~\cite{li2022grounded}, GroundingDINO~\cite{liu2024grounding}, and YOLO-World~\cite{cheng2024yolo} to localize different categories. 
Since for different categories, their parts are of different shapes and fine-grained features, it is hard to train universal detection or segmentation models.
For each image and given part labels, the detector is able to predict a set of bounding boxes in a zero-shot manner.
However, when dealing with less common animals (e.g., arthropods or mollusks) and parts (e.g., manes or antennae), these methods still struggle to achieve strong zero-shot performance.
Therefore, we perform fine-tuning with the annotated seed data, resulting in an improved part detection model. 
After obtaining the initial part detection results, we further utilize \texttt{GPT-4o mini} to check the part annotations in the form of VQA, filtering out errors in a more cost-effective and faster way.
Next, we apply the Segment Anything Model (SAM)~\cite{kirillov2023segment} to convert each bounding box $b \in B$ into a corresponding segmentation mask $M_b$ to obtain a more accurate representation.

\subsection{Multi-modal Knowledge Alignment and Completion}
\label{approach_align}
With extracted text-source knowledge entries, corresponding images, and region annotations, the final step is to associate the multi-modal knowledge together.
First, based on matching the category name, some images and regions can be directly mapped to the corresponding textual nodes.
Then, in the part hierarchy, there are many parts that are too trivial to be included in the knowledge text, which have regional annotations but lack corresponding text nodes to match. 
For such parts, the trivial part triplets “[animal]-Have-[part]” are inherited from the supercategories to the text knowledge entries to complete the knowledge base and are matched with the correct regions.
Besides, other similar entities are merged into the same node based on their BERT semantic similarity.
As a result, based on the alignment of visual entities, the region correspondence of visual triplets $\mathcal{R}^\text{vis}$ can be naturally regarded as a combination of the head entity and tail entity regions.
Finally, after alignment and completion, the triplet entries with their visual samples are validated by expert spot checks, forming a quality-assured multi-modal, multi-granular visual knowledge base.

\begin{figure}[t]
    \centering
    \includegraphics[width=\linewidth]{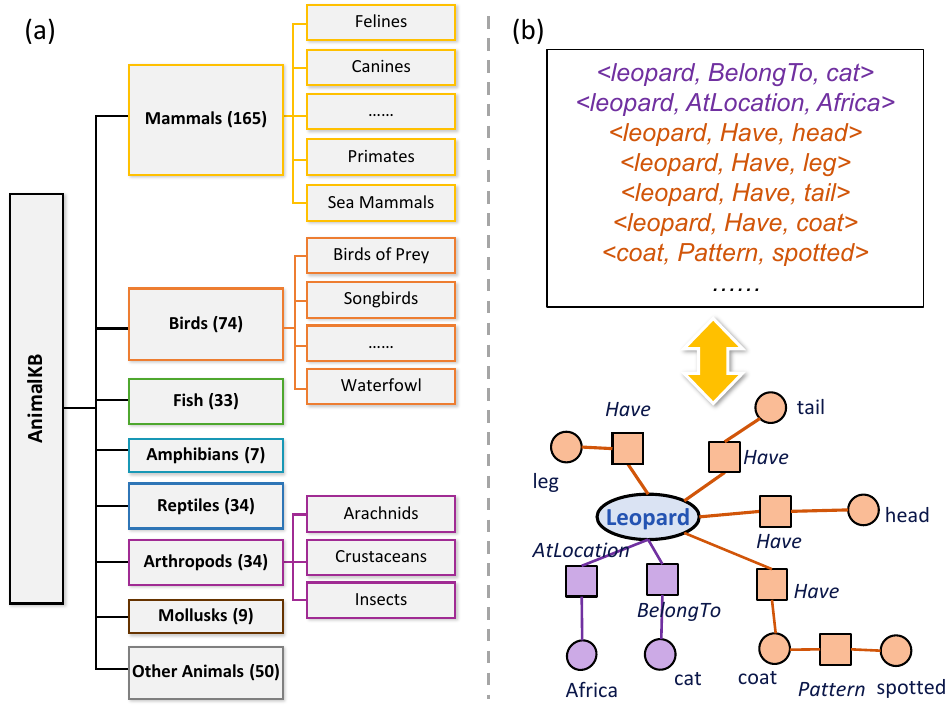}
    \caption{Visualization of animal category hierarchy and text knowledge examples. (a) Animal category hierarchical distribution. (b) Two equivalent forms of the textual knowledge examples.}
    \label{fig:animal_categories}
\end{figure}

\section{Data Statistics}
In this section, we present the key statistics about the constructed AnimalKB, compare it to the existing multi-modal and animal-specific visual knowledge bases, and provide typical visual examples in our knowledge base.

\begin{table*}[t]
    \centering
    \caption{Comparison with mainstream scene-, object-, and part-level visual datasets. AnimalKB is aimed at object-level visual understanding, involving a wide range of animal categories, a large number of images and region annotations, and provides multi-granular, multi-modal knowledge.}
    \label{tab:dataset_comp}
    \begin{tabular}{l|ccccccc}
        \toprule
        \textbf{Dataset}  & \textbf{Type} & \textbf{Category Num} & \textbf{Image Num} & \textbf{Object Annot.} & \textbf{Part Annot.} & \textbf{Knowledge}\\
        \midrule
        MSCOCO\cite{lin2014microsoft}        & scene-level  & 80 & 123,287 & $\checkmark$ & $\times$ & $\times$\\
        Visual Genome\cite{krishna2017visual} & scene-level  & 76,340 & 108,000 & $\checkmark$ & $\times$ & $\times$\\
        CUB-bird\cite{WahCUB_200_2011}      & object- and part-level & 200 & 11,788 & $\checkmark$ & $\checkmark$ & $\times$\\
        Stanford Dog\cite{Khosla2011StanfordDogs}  & object-level & 120 & 20,580 & $\checkmark$ & $\times$& $\times$ \\
        AwA2\cite{Xian2018ZeroShotLearning}          & object-level & 50 & 37,322 & $\times$ & $\times$ & $\times$\\
        iNat Dataset\cite{VanHorn2018iNaturalist}  & object-level & 5,089 & 859,000 & $\times$ & $\times$ & $\times$\\
        Pascal Part\cite{Chen2014DetectWhatWhere}   & part-level & 20 & 19,740 & $\checkmark$ & $\checkmark$ & $\times$\\
        % Our AnimalKB    & object- and part-level & 406 & 60,977/421,959 & $\checkmark$ & $\checkmark$ & $\checkmark$ \\
        Our AnimalKB    & object- and part-level & 406 & 421,959 & $\checkmark$ & $\checkmark$ & $\checkmark$ \\
        \bottomrule
    \end{tabular}
\end{table*}

\begin{figure}[t]
    \centering
    \includegraphics[width=\linewidth]{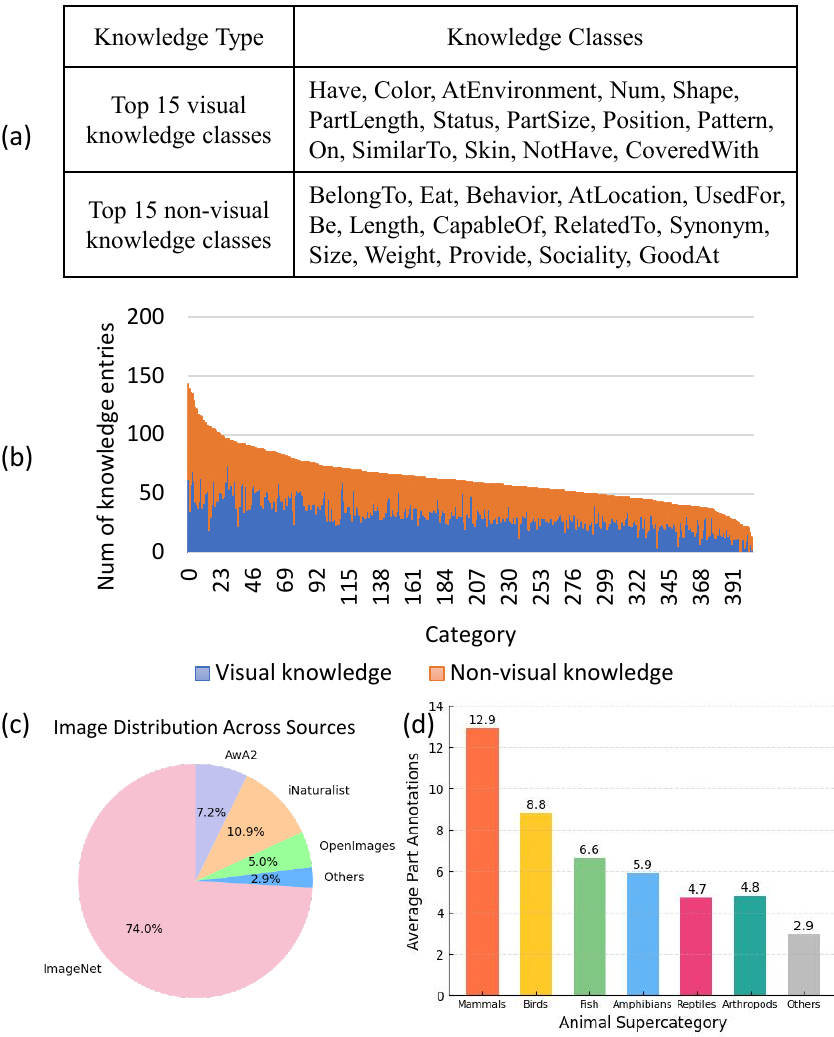}
    \caption{Statistical data of visual and non-visual knowledge. (a) Top 15 visual and non-visual knowledge classes. (b) Ratio of visual and non-visual knowledge triplets of each animal. (c) Image distribution over the sources. (d) Average number of part annotations for supercategories.}
    \label{fig:vis_non_vis_num2}
\end{figure}

\subsection{Text-source Knowledge Statistics}

We extract textual knowledge from 406 articles in the Encyclopedia Britannica for Kids, with an average article length of 27.3 sentences and 280.2 words. The knowledge base covers categories of popular science animals, including mammals, fish, birds, amphibians and reptiles, mollusks, arthropods, and others. 
The distribution of animal categories is shown in Fig.~\ref{fig:animal_categories}(a). 
Among these, mammals represent the largest supercategory, accounting for approximately 41\% of all animals, including 165 categories such as the Felidae, Canidae, Perissodactyla, Artiodactyla, and Marsupialia. 
Next in line are birds, reptiles, and arthropods, comprising 74, 34, and 34 categories, respectively. 
Note that the Encyclopedia Britannica Kids does not strictly adhere to a hierarchical structure based on biological taxonomy. Instead, it is divided into a relatively coarse three-level structure according to the cognitive habits of human children. We retain this hierarchical structure. 

In AnimalKB, textual knowledge is stored as triplets, as shown in Fig.~\ref{fig:animal_categories}(b), which is equivalent to the graph structure. AnimalKB contains 60 classes of knowledge (i.e. 60 relations), and categorizes them into visual and non-visual knowledge based on their observability in most image samples.
Fig.~\ref{fig:vis_non_vis_num2}(a) presents the top 15 visual and non-visual knowledge classes ranked by the number of triplets.
In total, 22,449 text knowledge entries are annotated, with 12,453 entries related to visual knowledge, accounting for approximately 55\%, and 9,996 entries related to non-visual knowledge, accounting for approximately 45\%. The proportion of visual knowledge and non-visual knowledge for each animal category is depicted in Fig.~\ref{fig:vis_non_vis_num2}(b).

\begin{figure*}[t]
\centering
\includegraphics[width=0.95\textwidth]{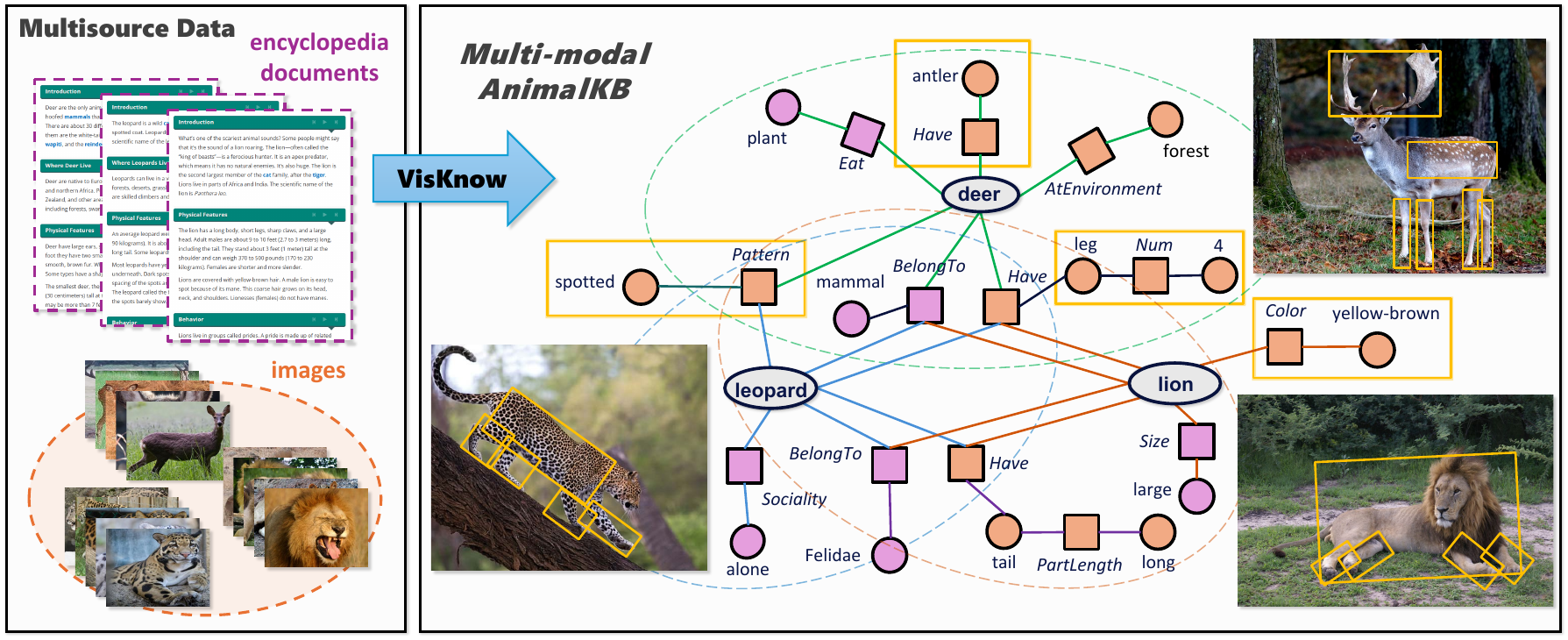}
\caption{Examples from the constructed AnimalKB. In the knowledge base, visual entities and relations correspond to images and regional annotations, which construct a more intuitive and comprehensive multi-granularity understanding of objects. Additionally, the nodes and relations might be shared among various categories, thereby establishing associations between animals and achieving inter-category understanding.}
\label{fig:kb_vis}
\end{figure*}

\subsection{Image-source Knowledge Statistics}
The triplets extracted from the text form the backbone of the knowledge base. Then, we expand the visual resources for the visual knowledge by associating corresponding images and image region annotations with the visual nodes and edges.

We construct visual resources for 340 out of 406 animal species in AnimalKB, while the remaining 66 categories of animals mainly include supercategories of animals and extinct animals.
For the images, we aggregate and collect them by category from existing datasets and the Internet, including AwA2, OpenImages, ImageNet, iNaturalist, and other sources, ensuring that each category contains at least 100 high-quality images. 
Altogether, AnimalKB contains 421,959 images.
The distribution of images across these sources is shown in the Fig.~\ref{fig:vis_non_vis_num2}(c).

In terms of part annotations, we first select up to 300 images for each animal category and perform automatic annotating on these samples.
In total, 94,626 images from AnimalKB are annotated with part regions.
On average, each animal category is annotated with 278.3 images, and each image is annotated with an average of 9.7 parts. Average part annotation for each animal supercategory are shown in Fig.~\ref{fig:vis_non_vis_num2}(d). More details are provided in the supplementary material.

Comparison with existing visual knowledge bases is presented in Tab.~\ref{tab:dataset_comp}. MSCOCO \cite{lin2014microsoft} and Visual Genome \cite{krishna2017visual} are general visual datasets but lack fine-grained, part-level annotations. Compared to specialized datasets focusing on specific animal categories like CUB \cite{WahCUB_200_2011} and Stanford Dog \cite{Khosla2011StanfordDogs}, we encompass a greater number of animal categories, more images, and more part annotations. While the iNaturalist Dataset \cite{VanHorn2018iNaturalist} has huge number of animal species and images, it lacks part annotations. Existing part annotation datasets like PascalPart \cite{Chen2014DetectWhatWhere} annotate parts in the form of masks manually, which is precise but resource-intensive, while our proposed framework strikes a balance between annotation accuracy and scalability. 
Furthermore, our visual regions are aligned with text knowledge entries, providing richer symbolic and semantic representations.

\subsection{Showcase of Animal Knowledge Base}
Fig.~\ref{fig:kb_vis} shows examples from AnimalKB. Visual knowledge includes annotations of animals and parts corresponding to text nodes, while non-visual knowledge consists of only text nodes. The same entities and knowledge triplets associate different animals, which is helpful for transfer learning among categories.

\section{KB-assisted Downstream Tasks}
\label{sec:app}

This section assesses how the constructed knowledge base, enriched with both textual and visual annotations, can effectively support downstream vision tasks that inherently demand external knowledge.
Rather than serving as mere auxiliary data in the single modality, the proposed AnimalKB is designed to bridge the gap between raw visual signals and high-level semantic understanding.
To demonstrate this, we select two representative and widely studied settings: zero-shot recognition (Sec. \ref{sec:zsl}), which probes the model's generalization ability across categories through multi-modal knowledge transfer, and fine-grained visual question answering (Sec. \ref{sec:vqa}), which stresses reasoning over subtle visual cues paired with knowledge-intensive attributes.
These tasks collectively highlight the wide range of challenges, such as generalization, reasoning, visual grounding, and fine-grained discrimination, that AnimalKB aims to address.
More details and corresponding prompts are provided in the supplementary material.

\subsection{Zero-shot Recognition}
\label{sec:zsl}

The process of adapting a recognition model from seen to unseen categories is formally defined as \textbf{zero-shot learning} (ZSL).
The incorporation of auxiliary external knowledge is typically required in ZSL to bridge the gap between seen and unseen categories.

\subsubsection{Task Description}

In recent years, the advent of 
vision-language models like CLIP~\cite{radford2021learning} has revolutionized zero-shot recognition.
These models are trained on massive datasets of image-text pairs, and learn a joint embedding space where images and text descriptions are aligned.
They perform zero-shot recognition by matching the visual feature of the input image with the textual features of candidate category labels, using the distance between the embeddings of category labels learned during pre-training as the implicit knowledge.
CLIP-based approaches have become the mainstream in zero-shot recognition due to their impressive generalization capabilities and ease of use.

In this setting, all categories in the datasets are treated uniformly as test categories, without distinguishing between seen and unseen ones.
Specifically, 340 leaf nodes (i.e. categories with image samples) from the category hierarchy of AnimalKB are designated for testing.
The goal of current recognition task is to accurately categorize a given test image into one of these 340 categories, with performance assessed by the final classification accuracy.

\begin{figure}[t]
\centering
\includegraphics[width=\linewidth]{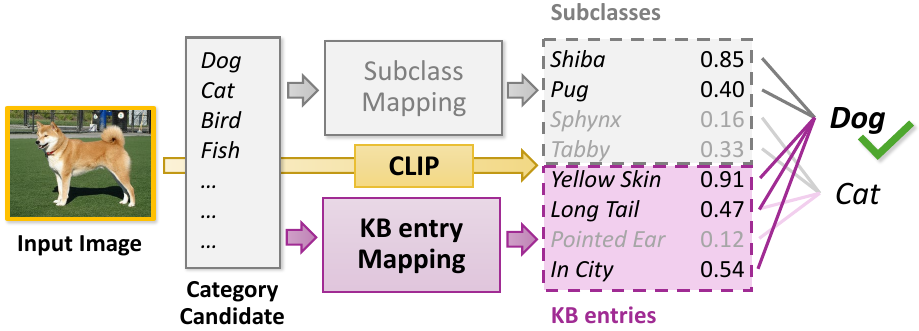}
\caption{Illustration of enhancing the CLIP's capacity of zero-shot fine-grained recognition by leveraging our constructed KB. To improve multi-modal matching, KB entries with semantically richer text descriptions transform the matching process from using fine-grained category names, which pre-trained multi-modal models may not handle well, to utilizing a set of common concepts. This enhances the matching through multi-concept integration.}
\label{fig:zsl}
\end{figure}

\subsubsection{Pipeline of KB-inside Zero-Shot Transfer}

Since the feature spaces for both images and text are fixed after the pre-training stage, the performance of CLIP-based ZSL methods is heavily influenced by the information richness of the textual descriptions related to the target categories (beyond category labels).
These descriptions serve as a bridge between the visual and semantic spaces, and their quality directly affects how well the model can align visual features with categories.
For example, a text description like \textit{a large African mammal with stripes} provides more discriminative information than a more general term like \textit{animal}, enabling the model to better distinguish within similar categories.
Therefore, the selection of textual descriptions becomes critical in this framework.

Prior works have addressed the challenge by enriching category descriptions, such as generating several subcategory labels to better align with test images~\cite{novack2023chils}.
However, these approaches are predominantly empirical and fail to offer targeted customization for specific categories.
We leverage the constructed AnimalKB to provide more comprehensive textual descriptions for each animal category, thereby further enhancing CLIP's zero-shot recognition capabilities, as illustrated in Fig. \ref{fig:zsl}.
Specifically, we transform the annotated triplets for each animal category (e.g., \textit{dog-AtEnvironment-in city} or \textit{cat-have-long tail}) into natural language phrases, serving as fine-grained descriptions of the target categories.
Similar to subcategory labels, these fine-grained descriptions offer additional semantic contributions to describe the categories with greater diversity.

To enhance the effectiveness of knowledge utilization, we preprocess the knowledge entries and employ an ensemble method to construct the final knowledge set.
Specifically, triplets that lack category discriminability, such as those inherited from parent categories or supplemented by common sense, are initially excluded.
Subsequently, adjacent triplets that can form concatenations are combined into a single description.
For instance, \textit{cat-Have-tail} and \textit{tail-PartLength-long} can be concatenated into \textit{cat-Have-tail-PartLength-long}, offering a more complete description for CLIP to capture semantics.
Finally, only the knowledge entries exhibiting the highest similarity to the visual features in the support set are selected to form the final set of entries.

\begin{table}[t]
    \centering
    \caption{Performance on zero-shot recognition with AnimalKB.}
    \label{tab:zsl}
    \begin{tabular}{lcc}
        \toprule
        \textbf{Setting} & \textbf{ViT-B/16} & \textbf{ViT-L/14} \\
        \midrule
        CLIP Baseline                        & 67.58 & 74.26 \\
        CLIP Subcategory                        & 69.38 & 76.13 \\
        CLIP Subcategory + Visual Knowledge     & 69.92 & 77.17 \\
        CLIP Subcategory + Non-visual Knowledge  & 70.33 & 77.32 \\
        CLIP Subcategory + All Knowledge        & \textbf{70.49} & \textbf{77.55} \\
        \bottomrule
    \end{tabular}
\end{table}

\begin{figure*}[t]
\centering
\includegraphics[width=\linewidth]{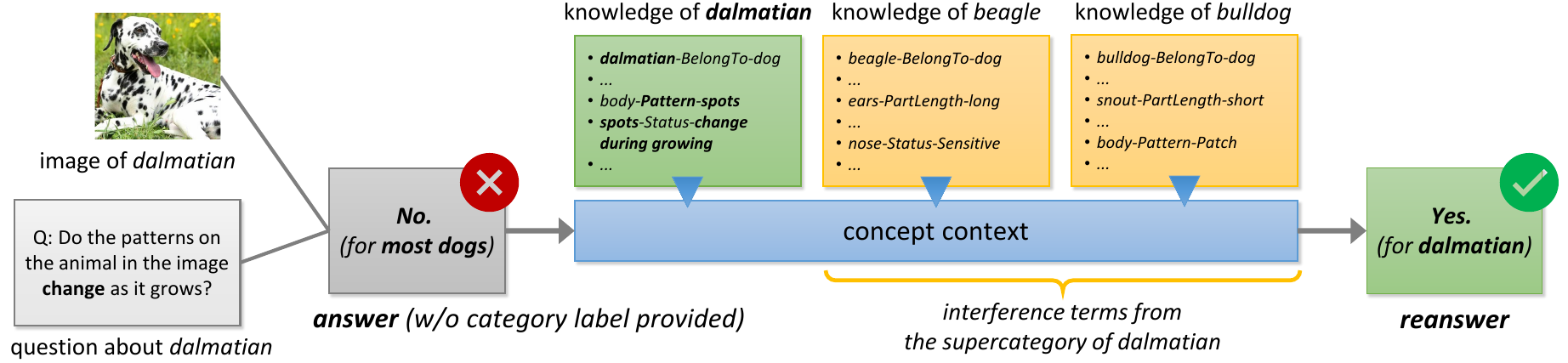}
\caption{Flow path of visual question answering with knowledge. \textit{Reanswering} is utilized to compare the impact of introducing knowledge on the model's ability to answer visual questions. Knowledge entries are provided to MLLMs as context, including interference terms, to simulate the noise knowledge typically present in practical applications of retrieval-augmented generation (RAG).}
\label{fig:vqa_flow}
\end{figure*}

\subsubsection{Experimental Results}

Tab.~\ref{tab:zsl} illustrates the zero-shot recognition performance of CLIP with various configurations.
The baseline performance of CLIP without any additional knowledge acts as a reference result.
The larger architecture, ViT-L/14, demonstrates superior performance due to its greater ability to capture complex patterns.
The inclusion of subcategory labels substantially improves performance.
This finding suggests that using only the category labels as the baseline does not sufficiently enable CLIP to match visual signals to precise fine-grained categories. Therefore, integrating subcategory information is highly beneficial in such settings.
Additionally, incorporating both visual and non-visual knowledge markedly enhances model performance.
The marginal advantage offered by non-visual knowledge can be attributed to the more frequent occurrence of non-visual text in the image-text pairs utilized during the CLIP pre-training process.
When all available knowledge is utilized, the model achieves optimal performance,
underscoring the synergistic effect of integrating multiple knowledge types.
These findings highlight the significance of integrating supplementary knowledge into the CLIP framework for zero-shot recognition tasks.

In conclusion, the diversity and richness of category-related text descriptions (considered as various ``semantic concepts'' that describe the categories) enhance the robustness of image-text matches.
This robustness is vital for the effective transferability of pre-trained multi-modal models in practical applications.
Our constructed KB offers the precise, high-quality semantic concepts required in these contexts, demonstrating a successful practice in the application of multi-modal methods.
These results offer valuable insights for future research focusing on the incorporation of structured knowledge.

\subsection{Fine-grained VQA}
\label{sec:vqa}

Visual Question Answering (VQA) is a prominent task in computer vision, serving as the most intuitive interface for visual models to interact with human users.
While the rapid advancement of multi-modal pre-trained models has led to significant progress in general-domain QA, these models continue to encounter challenges in specialized domains.

\subsubsection{Task Description}

We assess the VQA capabilities of both closed-source and open-source pre-trained multi-modal models within the animal domain.
The models answer identical questions based on input images under two conditions: ``without KB'' and ``with KB'' to evaluate their Retrieval-Augmented Generation (RAG) \cite{gupta2024comprehensive,yasunaga2022retrieval,guu2020retrieval,caffagni2024wiki} capabilities.

We develop a comprehensive VQA benchmark dataset using AnimalKB, where each question is tied to an image of a specific animal category.
The questions are automatically generated using GPT-4o from category-level textual triplets.
We instruct GPT-4o to avoid including animal names in the questions to ensure that the tested model does not rely solely on text, thereby necessitating the use of images.
The triplets generate two types of questions based on their characteristics: visual and non-visual.
Visual questions evaluate the model's ability to process images, permitting it to respond without identifying the animal's category.
Non-visual questions assess the model's overall competence, requiring it to first recognize the animal's category from the image, and then answer based on the model's embedded world knowledge.
Furthermore, for each animal category, a support set is made from the textual triplets to provide reference information, aiming to evaluate whether the model can enhance its accuracy with the assistance of KB.

\begin{table}[t]
    \centering
    \caption{Performance of closed-source MLLMs on VQA with AnimalKB.}
    \label{tab:vqa_closed_source}
    \begin{tabular}{lccc}
        \toprule
        \textbf{Model} & \textbf{Visual} & \textbf{Non-visual} & \textbf{All} \\
        \midrule
        \multicolumn{4}{l}{\textbf{without KB}} \\
        gpt-4o (1120)       & 78.95 & 76.81 & 77.88 \\
        gpt-4o (0806)       & 79.36 & 77.13 & 78.25 \\
        gpt-4o-mini         & 69.78 & 66.86 & 68.32 \\
        gemini-2.0-flash    & \underline{\textbf{84.95}} & \underline{\textbf{81.96}} & \underline{\textbf{83.46}} \\
        gemini-1.5-pro      & 78.68 & 78.68 & 79.71 \\
        % gemini-1.5-flash    & 71.42 & 70.22 & 70.82 \\
        claude-3.5-sonnet   & 81.05 & 77.82 & 79.44 \\
        % claude-3.5-haiku    & & & \\
        claude-3-haiku      & 63.82 & 59.14 & 61.48 \\
        \midrule
        \multicolumn{4}{l}{\textbf{with KB}} \\
        gpt-4o (1120)       & 98.26 & 97.87 & 98.06 \\
        gpt-4o (0806)       & \underline{\textbf{98.68}} & \underline{\textbf{97.79}} & \underline{\textbf{98.24}} \\
        gpt-4o-mini         & 94.04 & 95.07 & 94.56 \\
        gemini-2.0-flash    & 98.16 & 97.65 & 97.90 \\
        gemini-1.5-pro      & 97.01 & 96.54 & 96.78 \\
        % gemini-1.5-flash    & 96.89 & 97.06 & 96.97 \\
        claude-3.5-sonnet   & 97.43 & 96.72 & 97.07 \\
        % claude-3.5-haiku    & & & \\
        claude-3-haiku      & 92.06 & 92.21 & 92.13 \\
        \bottomrule
    \end{tabular}
\end{table}

\begin{table}[t]
    \centering
    \caption{Performance of open-source MLLMs on VQA with AnimalKB.}
    \label{tab:vqa_open_source}
    \begin{tabular}{llccc}
        \toprule
        \textbf{Model} & \textbf{Size} & \textbf{Visual} & \textbf{Non-visual} & \textbf{All} \\
        \midrule
        \multicolumn{4}{l}{\textbf{without KB}} \\
        DeepSeek-VL2 & 27.5 B & 73.75 & \underline{\textbf{75.69}} & \underline{\textbf{74.72}} \\
        DeepSeek-VL2-Small & 16.1 B & 66.13 & 68.70 & 67.41 \\
        % DeepSeek-VL2-Tiny & 3.4 B & 64.02 & 62.06 & 63.04 \\
        Qwen2.5-VL-7B & 8.3 B & 74.44 & 71.25 & 72.84 \\
        Qwen2-VL-7B & 8.3 B & 74.56 & 72.84 & 73.70 \\
        InternVL2.5-8B & 8.1 B & 71.27 & 69.09 & 70.18 \\
        InternVL2.5-26B & 25.5 B & \underline{\textbf{76.52}} & 72.55 & 74.53 \\
        % InternVL2-8B & 8.0 B & 71.08 & 68.97 & 70.02 \\
        % InternVL2-26B & 25.5 B & 72.61 & 69.03 & 70.82 \\
        % LLaVA-v1.6-Mistral-7B & 7.6 B & 65.54 & 60.10 & 62.82 \\
        LLaVA-v1.6-Vicuna-7B & 7.1 B & 65.76 & 60.91 & 63.33 \\
        % LLaVA-v1.6-Vicuna-13B & 13.4 B & 64.75 & 62.70 & 63.73 \\
        LLaVA-Next-Llama3-8B & 8.4 B & 65.47 & 63.75 & 64.61 \\
        LLaVA-OV-Qwen2-7B & 8.0 B & 71.94 & 65.27 & 68.60 \\
        \midrule
        \multicolumn{4}{l}{\textbf{with KB}} \\
        DeepSeek-VL2 & 27.5 B & 84.66 & 85.10 & 84.88 \\
        DeepSeek-VL2-Small & 16.1 B & 85.39 & 84.73 & 85.06 \\
        % DeepSeek-VL2-Tiny & 3.4 B & 71.96 & 72.84 & 72.40 \\
        Qwen2.5-VL-7B & 8.3 B & 87.06 & 85.42 & 86.24 \\
        Qwen2-VL-7B & 8.3 B & 87.67 & 86.91 & 87.29 \\
        InternVL2.5-8B & 8.1 B & 87.23 & 86.47 & 86.85 \\
        InternVL2.5-26B & 25.5 B & \underline{\textbf{87.99}} & \underline{\textbf{88.04}} & \underline{\textbf{88.01}} \\
        % InternVL2-8B & 8.0 B & 86.27 & 85.20 & 85.74 \\
        % InternVL2-26B & 25.5 B & 87.16 & 85.29 & 86.23 \\
        % LLaVA-v1.6-Mistral-7B & 7.6 B & 82.11 & 81.81 & 81.96 \\
        LLaVA-v1.6-Vicuna-7B & 7.1 B & 80.71 & 80.86 & 80.78 \\
        % LLaVA-v1.6-Vicuna-13B & 13.4 B & 82.13 & 81.74 & 81.94 \\
        LLaVA-Next-Llama3-8B & 8.4 B & 83.41 & 84.19 & 83.80 \\
        LLaVA-OV-Qwen2-7B & 8.0 B & 83.28 & 81.81 & 82.55 \\
        \bottomrule
    \end{tabular}
\end{table}

\subsubsection{Evaluation Metrics}

For each question in the QA set, we employ a two-stage evaluation process as depicted in Fig.~\ref{fig:vqa_flow}.
Initially, the model is prompted to generate an \textbf{answer} based on the image.
Subsequently, it is prompted to \textbf{re-answer} using the initial response supplemented by additional knowledge.
To eliminate the impact of retrieval models, we provide one set of accurate, relevant knowledge and three sets of irrelevant, distracting knowledge for each question presented in a random order, thereby simulating a consistent condition of an RAG process.
We calculate the accuracy of the answers at each stage separately.
The first stage assesses the model's capability for fine-grained QA, whereas the second stage evaluates its proficiency in leveraging extra information.
Furthermore, we categorize the questions and knowledge into ``visual'' and ``non-visual'' types to analyze the accuracy of answers according to question type and the effectiveness of different types of knowledge.

\subsubsection{Experimental Results}

The performance analysis of both closed-source (\cite{openai2024gpt4o,pichai2024gemini2,anthropic2024claude35sonnet}, Tab.~\ref{tab:vqa_closed_source}) and open-source (\cite{wu2024deepseekvl2,bai2025qwen25vl,bai2024qwen2vl,internvl2024internvl25,internvl2024internvl2,liu2024llavav16}, Tab.~\ref{tab:vqa_open_source}) MLLMs reveals several key insights.

The integration of KB significantly enhances the performance of models, irrespective of their architecture or size.
For example, in Tab.~\ref{tab:vqa_closed_source}, the accuracy of \texttt{gpt-4o} increases markedly by 20.18\% with the utilization of KB.
Similarly, in Tab.~\ref{tab:vqa_open_source}, the accuracy of \texttt{InternVL2.5-26B} improves by 13.48\% upon the inclusion of KB.
This consistent trend across all models demonstrates that KB integration is an effective choice in enhancing model performance.

The size of the model significantly influences its performance, especially within the open-source category.
This suggests that larger models possess an enhanced capacity for processing and integrating complex information, resulting in superior performance.
However, even models of similar scale can have different performance, and their ability to answer questions with the KB also varies.

In conclusion, the evaluation of both closed-source and open-source MLLMs demonstrates the significant impact of our constructed KB on model performance.
The integration of the KB consistently enhances accuracy across all models, with notable improvements observed in both closed-source models such as \texttt{gpt-4o} and open-source models such as \texttt{InternVL2.5-26B}.
These insights highlight the critical role of integrating KBs to enhance the effectiveness of MLLMs within specialized domains.

\subsection{Summary}

The results from the experiments indicate that the constructed KB can enhance model performance for various visual tasks.
Notably, for large multi-modal pre-trained models with already strong performance, incorporating domain-specific expert knowledge aids in better contextual understanding, thereby making these models more reliable than relying solely on their implicit pre-training knowledge.
While the experiments show that our KB can effectively support visual models, our current use of the KB is still in its preliminary phase.
We anticipate that future research will yield greater benefits by more thoroughly utilizing the knowledge within the KB.

\section{KB as benchmark}
\label{sec:bench}

AnimalKB's dual-modality structure, integrating textual triplets with part-level visual-text alignment, not only supports downstream models but also serves as a rigorous benchmark for evaluating knowledge reasoning and detailed visual understanding across both modalities.
Its explicit semantics and comprehensive part annotations present challenges that surpass those in conventional benchmarks.
Besides, it can be used to evaluate the models' generalization and transfer capabilities on encyclopedia-style data of animal domain.
In this section, we demonstrate its benchmark value through two complementary tasks: knowledge graph completion (Sec.~\ref{sec:textbench}), which assesses the ability of NLP models to infer missing (particularly vision-related) entities, and part segmentation (Sec.~\ref{sec:partbench}), which evaluates the ability of vision models to achieve precise part-level understanding.
Collectively, these tasks illustrate how AnimalKB provides a comprehensive and challenging testbed on animal domain, encompassing both symbolic reasoning and structured visual perception.

\subsection{TextBench}
\label{sec:textbench}

\subsubsection{Task Description}

Using the textual knowledge triplets in \textbf{AnimalKB}, we evaluate how well existing knowledge graph models perform reasoning and completion on encyclopedic, animal-domain data, known as the task of Knowledge Graph Completion (KGC). 
Specifically, for a triplet $(h, r, t)$, the task is to predict the missing entity in head- and tail-prediction queries of the form $(?, r, t)$ and $(h, r, ?)$, respectively.
The benchmark consists of the textual triplets of AnimalKB where the length of the head and tail entities are both less than or equal to three. The dataset is split into training and testing sets with an 85\%:15\% ratio.

\subsubsection{Evaluation Metrics}

We evaluate the performance of the models using two standard metrics in knowledge graph reasoning tasks.

\begin{itemize}
\item \textbf{MRR (Mean Reciprocal Ranking)} computes the average of the reciprocal ranks of the predicted entities for all test triplets. Higher MRR means better performance.
\item \textbf{HITS@k} calculates the proportion of test triplets where the correct entity is ranked within the top-10 predictions. Higher HITS@k indicates better performance.
\end{itemize}

\begin{table}[t]
\centering
\caption{Performance of models on the knowledge graph completion task.}
\label{tab:kgc_performance}
\begin{tabular}{lcccc}
\toprule
\textbf{Model} & \textbf{MRR} & \textbf{HITS@1} & \textbf{HITS@3} & \textbf{HITS@10} \\
\midrule
\multicolumn{5}{l}{\textbf{Embedding-based methods}} \\
TransE~\cite{bordes2013translating}          & 18.1  & 11.5 & 22.0 & 29.8    \\
ComplEx~\cite{trouillon2016complex}         & 12.0  & 10.7 & 12.2 & 14.7     \\
DistMult~\cite{yang2014embedding}        & 11.0 & 9.2 & 11.3 & 14.1      \\
RotatE~\cite{sun2019rotate}          & 19.6  & 15.8 & 21.4 & 26.5      \\
\midrule
\multicolumn{5}{l}{\textbf{Text-based methods}} \\
KG-Bert~\cite{yao2019kg}        & 21.1  & 12.3  & 23.9 & 38.2    \\
StAR~\cite{li2022star}            & 30.9  & 23.6 & 33.7 & 44.6    \\ 
SimKGC~\cite{wang2022simkgc}       & \textbf{38.6}  & \textbf{32.4} & \textbf{40.7} & \textbf{50.2}    \\
\midrule
\multicolumn{5}{l}{\textbf{LLM-based methods}} \\ 
gpt-4o (0806) & 42.9  & \textbf{34.3} & 46.0 & 60.8\\
gemini-1.5-pro & \textbf{44.2}  & 28.8 & \textbf{52.3} & \textbf{81.1} \\ 
\bottomrule
\end{tabular}
\end{table}

\subsubsection{Experimental Results}

Tab.~\ref{tab:kgc_performance} presents the performance of various KGC models across three groups: embedding-based methods, text-based methods, and LLM-based methods.
The former two groups require models to be trained on the training set before evaluation, whereas LLM-based methods can be directly evaluated without task-specific training by simply providing in-prompt examples and explicit output-format instructions.

Among the embedding-based methods, \texttt{RotatE} achieves the best performance, demonstrating its superiority in capturing the relational structures within the knowledge graph compared to other embedding-based models like \texttt{TransE}, \texttt{ComplEx}, and \texttt{DistMult}. Notably, \texttt{TransE} achieves a competitive HITS@10 of 29.8, slightly outperforming \texttt{RotatE} when the candidate entity range is relatively broad.
For the text-based methods, \texttt{SimKGC} exhibits the strongest performance, with an MRR of 38.6 and a HITS@10 of 50.2, surpassing both \texttt{KG-BERT} and \texttt{StAR}, which suggests its effectiveness in leveraging textual context for knowledge graph completion. 

LLM-based methods outperform both embedding-based and text-based approaches with only few examples displayed in the prompt. The \texttt{gpt-4o} model achieves an MRR of 42.9 and a HITS@1 of 34.3, better than \texttt{SimKGC}. While the \texttt{gemini-1.5-pro} model performs worse on HITS@1, it demonstrates better performance on other metrics, achieving an MRR of 44.2 and a HITS@10 of 81.1. 
This indicates that the overall performance of \texttt{gemini-1.5-pro} is relatively high, but it is not accurate enough when distinguishing entities that are closest to the ground truth.

In summary, the results indicate that while traditional embedding and text-based methods have their strengths, such as smaller model sizes and faster speeds, LLM-based methods currently represent the state-of-the-art performance in knowledge graph completion in a few-shot manner and can be more easily transferred to data from different domains.

\subsection{PartBench}
\label{sec:partbench}

\begin{table}[t]
    \centering
    \caption{Performance of models on the instance segmentation task.}
    \label{tab:inst_seg_res}
    \resizebox{\columnwidth}{!}{
    \begin{tabular}{l c c c c c c c c}
        \toprule
        Model & Finetuned & Total AP & AP50 & AP75 & head & torso & leg \\
        \midrule
        VLPart & $\times$     & 5.09  & 8.86  & 5.23  & 18.12 & 4.33 & 4.79 \\
               & \checkmark   & 16.07 & 27.38 & 16.70 & 27.39 & 24.32 & 5.84 \\
        \midrule
        PartGLEE & $\times$     & 14.50 & 25.96 & 14.28 & 43.16 & 17.91 & 19.33 \\
                 & \checkmark  & \textbf{31.98} & \textbf{50.63} & \textbf{32.78} & \textbf{55.21} & \textbf{41.36} & \textbf{31.01} \\
        \bottomrule
    \end{tabular}
    }
\end{table}

\begin{table}[t]
    \centering
    \caption{Performance of models on the semantic segmentation task.}
    \label{tab:sem_seg_res}
    \resizebox{\columnwidth}{!}{
    \begin{tabular}{l c c c c c c c}
        \toprule
        Model & Finetuned & mIoU & fwIoU & head & torso & leg\\
        \midrule
        ClipSeg & $\times$     & 7.47  & 16.16 & 35.38 & 0.26 & 40.72 \\
                & \checkmark  & 14.40 & 37.61 & \textbf{55.87} & 52.74 & 42.41\\
        \midrule
        PartGLEE & $\times$     & 24.24 & 37.97 & 55.84  & 28.78 & 56.44\\
                 & \checkmark  & \textbf{36.87} & \textbf{44.96 }& 48.62  & 51.91 & \textbf{62.29}\\
        \midrule
        VLPart$_{sem}$ & $\times$    & 7.43  & 9.19  & 15.77 & 4.35 & 7.50\\
                    & \checkmark & 25.77 & 42.49 & 45.98 & \textbf{58.46} & 53.53\\
        \bottomrule
    \end{tabular}
    }
\end{table}

\subsubsection{Task Description}

In this experiment, we aim to evaluate various part segmentation models utilizing the part annotations of our AnimalKB. 
Since AnimalKB provides masks that distinguish different instances of each part, it can naturally be used to evaluate instance segmentation. 
Then, we merge the masks of all instances belonging to the same part to evaluate semantic segmentation.
We construct a subset that comprises 41 categories of mammals that are overlapped with the AwA2 dataset with annotations of 24 different parts. The subset is divided into training and testing splits with an 80\%:20\% ratio, where the training set contains 17,161 images with 225,371 annotated parts, and the test set contains 4,310 images with 56,615 parts.

\subsubsection{Evaluation Metrics}

Since semantic segmentation focuses on pixel-level category consistency while instance segmentation emphasizes instance-level detection and discrimination, we adopt IoU-based and AP-based metrics for evaluating the two tasks, respectively, following their standard evaluation protocols. 

For semantic segmentation, we calculate IoU of each part and evaluate \textbf{Mean IoU (mIoU)} and \textbf{Frequency Weighted IoU (fwIoU)}, where IoU is the intersection-over-union ratio between the predicted and ground truth areas. mIoU averages IoU across all classes to assess overall segmentation quality, while fwIoU weights each class IoU by its pixel frequency to better reflect performance under class imbalance. 

For instance segmentation, we calculate AP of each part and evaluate \textbf{Total AP}, \textbf{AP50}, \textbf{AP75}, where AP is computed as the area under the precision-recall curve across varying confidence thresholds. Total AP averages AP over multiple IoU thresholds to evaluate overall instance-level performance, with AP50 and AP75 calculating AP at IoU thresholds of 0.50 and 0.75, respectively, where a higher threshold represents a stricter matching criterion.

\subsubsection{Experimental Results}

We conduct experiments using three part segmentation models: \texttt{VLPart}~\cite{zhang2023vlpart}, \texttt{PartGLEE}~\cite{li2024partglee}, and \texttt{ClipSeg}~\cite{luddecke2022clipseg}, across both instance segmentation and semantic segmentation tasks. 
Among them, \texttt{ClipSeg} can only perform semantic segmentation, while \texttt{PartGLEE} supports both tasks. \texttt{VLPart} supports instance segmentation, and we merge the masks of the same part for evaluating semantic segmentation, denoted as \texttt{VLPart}$_{sem}$.
The experimental results presented in Tab.~\ref{tab:inst_seg_res} and \ref{tab:sem_seg_res} provide a comparison of model performance before and after fine-tuning with the AnimalKB data.

Tab.~\ref{tab:inst_seg_res} presents the results of the instance segmentation task. 
It can be seen that \texttt{PartGLEE} has a significant performance advantage compared to \texttt{VLPart}.
Fine-tuning brings large improvements to both models, with Total AP nearly doubling, indicating that it is an effective approach for domain transfer.
Similar results are observed for semantic segmentation, as shown in Tab.~\ref{tab:sem_seg_res}, where \texttt{PartGLEE} achieves the best overall performance and fine-tuning benefits all models.
In addition, due to the bias in the pre-training data, models before finetuning exhibit large performance discrepancies across different parts.
After fine-tuning on the AnimalKB data, harder parts such as the torso receive more improvements, which effectively narrows these performance gaps.

\begin{figure}[t]
    \centering
    \includegraphics[width=0.95\linewidth]{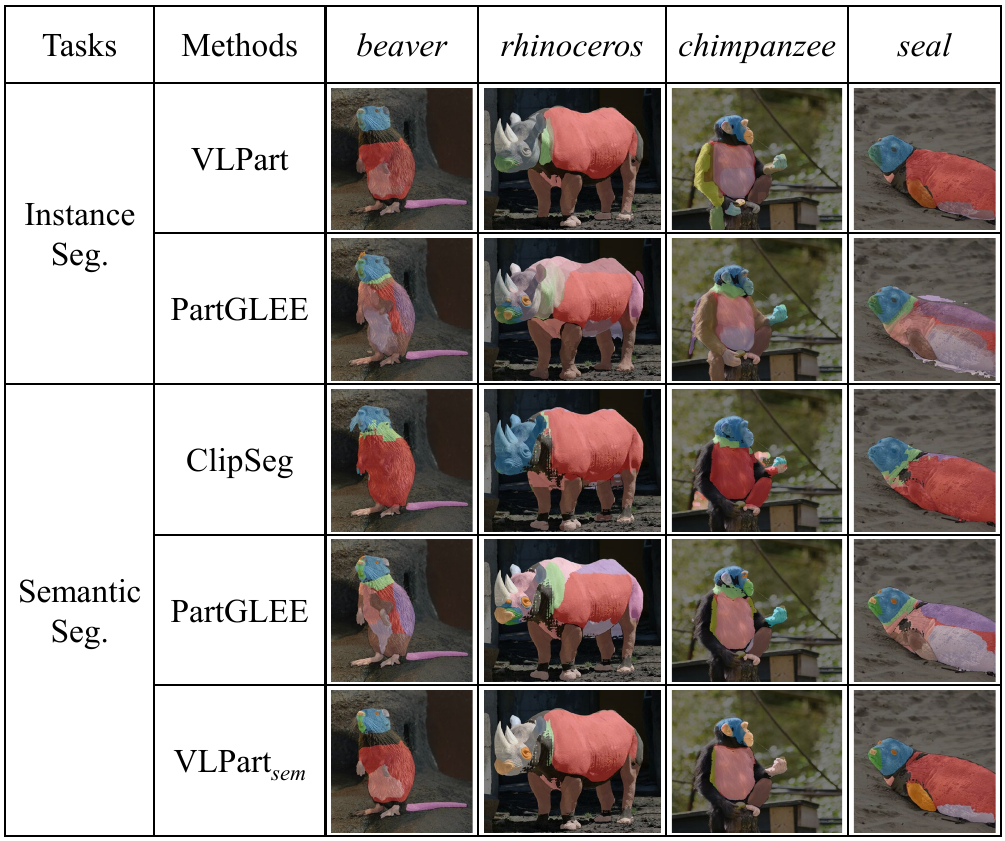}
    \caption{Visualization of segmentation results of various models. The constructed AnimalKB serves as a more comprehensive and precise benchmark for animal part segmentation, thus facilitating a more thorough evaluation of the effectiveness of part segmentation methods.}
    \label{fig:parts}
\end{figure}

\begin{figure*}[t]
    \centering
    \includegraphics[width=0.85\textwidth]{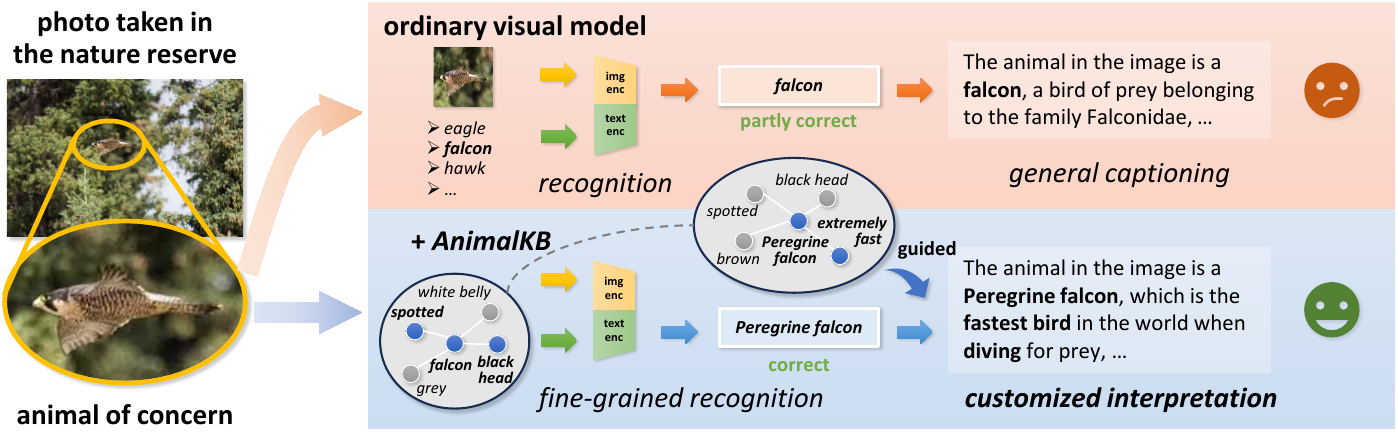}
    \caption{Illustration of the case study on KB application in natural reserves. Integrating the knowledge base enables an efficient pipeline including fine-grained recognition and customized interpretation of animal species in nature reserve photographs, thereby providing more accurate and contextually relevant results than general visual models in practical applications.}
    \label{fig:case_study}
\end{figure*}

Examples of segmentation results are shown in Fig.~\ref{fig:parts}. 
Overall, instance recognition models can cover a broader range of parts compared to semantic recognition, and it possesses the ability to distinguish between different components of the same category, such as identifying the individual legs of an animal. 
Among the various models, \texttt{PartGLEE} achieves the best performance on both tasks, consistent with the quantitative experimental results.
Furthermore, models perform better on quadruped animals like beavers and rhinoceroses compared to bipedal animals like chimpanzees and aquatic animals like seals. This is because their previous pre-training data predominantly consists of quadruped mammals, while our AnimalKB offers more challenging tests on animals with more distinct visual differences.

Overall, AnimalKB can provide challenging benchmarks on part-level perception for evaluating existing models in the animal domain, such as semantic segmentation and instance segmentation tasks.
Meanwhile, fine-tuning with AnimalKB consistently enhances model performance on both tasks, confirming the dataset’s effectiveness.
While certain part-level metrics still exhibit room for improvement, the results collectively highlight AnimalKB’s value in advancing fine-grained part segmentation.

\subsection{Summary}
The above experiments take knowledge graph completion and part segmentation as examples, validating that the proposed AnimalKB can serve as novel challenging benchmarks to for existing tasks. 
Compared with traditional unimodal benchmarks (which only use text or images), AnimalKB offers multi-modal aligned annotated data, enabling models to jointly learn from multiple modalities, and effectively test model performance and stability in handling complex multi-modal data.
Moreover, AnimalKB's fine-grained animal encyclopedia-style data introduces a certain domain gap with existing datasets, enabling evaluation of model generalization in this specialized domain.
Fine-tuning on AnimalKB training data can significantly improve model performance in the animal domain, helping the model better understand and reason about animal-related knowledge, making it an effective approach for cross-domain knowledge transfer.

In conclusion, AnimalKB provides a comprehensive and challenging testing platform for existing tasks in the animal domain, encouraging models to engage in multi-modal learning and domain generalization. Its rich variety of annotated data holds the potential to be expanded as benchmarks for a wider range of text, visual, and multi-modal tasks.

\section{Case Study: KB-guided Interpretation}
\label{sec:casestudy}

In this case study, we explore an application tailored for fine-grained visual captioning, with a focus on identifying and interpreting animals in environments such as zoos and nature reserves.
The challenge stems from distinguishing between fine-grained animal categories
and delivering descriptions that are engaging, informative, and genuinely meaningful to the audience.
To address this, we incorporate the proposed AnimalKB to augment the model's recognition and captioning capabilities, thereby ensuring precise identification and generating informative captions that cater to diverse user requirements, as illustrated in Fig.~\ref{fig:case_study}.

As introduced in Sec.~\ref{sec:zsl},
we enhance the pre-trained CLIP model by integrating it with AnimalKB that contains detailed information about animal species, including taxonomic hierarchies, habitat specifics, and distinguishing characteristics.
The knowledge base serves as an auxiliary information source, allowing the model to refine its predictions by cross-referencing visual attributes with domain-specific knowledge.
For instance, if the model detects a bird with a particular wing pattern, it can query the KB to narrow down the possible species based on additional attributes, such as geographic distribution or behavioral traits.
This hybrid approach could substantially improve the model's zero-shot recognition accuracy, ensuring the correct identification of rare species.

Once the accurate category is identified, the application utilizes the KB to generate informative and contextually relevant captions.
This captioning process is tailored to accommodate the audience's age, knowledge level, and specific interests~\cite{t4}.
For example, a child visitor might be provided with a simple, engaging description that highlights the animal's color and behavior, whereas a biology student might receive a more detailed explanation, including scientific names, evolutionary traits, and ecological significance.
The KB allows for customization based on user preferences, such as emphasizing conservation status or cultural importance, ensuring the captions are both educational and engaging.

This case study reveals how integrating a domain-specific knowledge base with a vision–language model enables both precise fine-grained recognition and audience-aware interpretation.
By leveraging structured animal knowledge, the system delivers captions that are not only accurate but also contextually meaningful for diverse users.
These results highlight the potential of hybrid knowledge-driven approaches in creating more informative, adaptable, and educational visual understanding applications.

\section{Conclusion}

In this work, we introduce \textbf{VisKnow}, a framework for constructing multi-modal, object-level knowledge graph that supports in-depth object understanding.
Using common animal categories as a representative domain, we develop \textbf{AnimalKB}, a knowledge base that integrates and aligns textual and visual information at both the object and part levels.
Experimental results show that the constructed KB effectively enhances zero-shot visual recognition and improves the knowledge question-answering capabilities of vision–language models.
Beyond these improvements, the richly annotated and fine-grained data in AnimalKB enables the creation of new, challenging benchmark datasets for tasks such as knowledge reasoning and part segmentation, while also demonstrating strong potential in downstream applications like fine-grained image description across expert knowledge domains.
We hope that this work provides valuable insights into visual knowledge base construction and utilization, and contributes to advancing deeper object understanding in practical multi-modal applications.

\section*{Acknowledgments}

This work is partially supported by Natural Science Foundation of China under contract No. U21B2025, and National Key R\&D Program of China No. 2023YFF1105104, 2021ZD0111901.

\ifCLASSOPTIONcaptionsoff
  \newpage
\fi

\bibliography{main_processed}
\bibliographystyle{IEEEtran}

\clearpage



\section*{Supplementary material}

# Supplementary Materials for
# VisKnow: Constructing Visual Knowledge Base
# for Object Understanding

Ziwei Yao, *Student Member, IEEE*, Qiyang Wan, *Student Member, IEEE*,
Ruiping Wang*, *Senior Member, IEEE*, and Xilin Chen, *Fellow, IEEE*

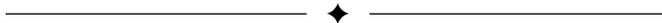

In this supplementary material, we provide more details of the constructed AnimalKB in Sec. 1, and some visualization examples of knowledge in AnimalKB are illustrated in Sec. 2. Next, we present more experimental details and discussions on downstream tasks in Sec. 3, and the prompts used in both the construction pipeline and the applications discussed in the main paper in Sec. 4. Additionally, we provide user interfaces of the knowledge base annotation and exploration systems in Sec. 5.

## 1 DETAILS OF ANIMALKB

In this section, we provide more details of our constructed AnimalKB, including some visual examples in Fig. 1 and a summary table of categories and parts in Tab. 1.

To illustrate the comprehensive and fine-grained nature of the constructed AnimalKB, we provide a selection of visual examples for the part-level annotations in Fig. 1. This figure demonstrates the annotation quality and the challenges successfully addressed across a few representative species (deer, horse, leopard, elephant, giraffe, chimpanzee, seal, dog, polar bear, and pig) and their distinct body parts, ranging from general features like the head and torso to highly specific ones like the foot and tail. Each row corresponds to a specific part (e.g., head, torso, front leg), and each column represents a different animal species. Grey cells indicate cases where the part is not visible or not relevant to the species.

Crucially, the annotations maintain high precision despite the dramatic morphological differences (e.g., comparing the neck of a giraffe to that of a seal) and varying poses (e.g., the front leg in a standing deer versus a sitting dog). The consistent and precise localization using yellow bounding boxes, even for partially obscured or small parts, confirms the dataset's suitability for tasks requiring fine-grained visual understanding, such as part-based recognition. These examples validate that our annotation protocol

effectively handles the complexity and diversity of real-world animal imagery.

To provide a comprehensive overview of the structure and coverage of AnimalKB, Tab. 1 presents the category-level details for the included species. The table is structured hierarchically, grouping entries by major group (e.g., Amphibians, Arthropods) and subgroup (e.g., Arachnids, Crustaceans, Insects), and then listing the specific animal category (e.g., african bullfrog, tarantula, bee). Critically, it specifies the annotated body parts for each category, demonstrating the fine-grained level of detail, which includes parts like the head, eye, foot for most categories, and the beak, wing for birds, and so on. Furthermore, the table quantifies the dataset's depth by showing the number of the text-source triplets collected (#triplets), the total number of annotated images (#annot. images) for each category, and the average number of bounding boxes (#avg. bboxes) present in each image.

It is important to note that because our knowledge base is derived from child-oriented encyclopedias, the resulting taxonomic hierarchy does not strictly adhere to formal biological classification. Instead, it employs a structure more widely familiar to the general public. Nevertheless, even under this less rigorous framework, the effective transfer of descriptive triplets from parent categories to subclasses remains achievable, which demonstrates a key advantage of utilizing such encyclopedia-type knowledge sources. This point can be observed in the "parts" section of the table, where the finer-grained animal categories always share the parts of their parent categories, and other triplets within the knowledge base exhibit similar characteristics. This increases the density of our knowledge base and enhances the connectivity between parent and child categories, which is beneficial for applying a more robust transferability during subsequent knowledge base utilization.

As a result, this meticulous categorization and quantification confirm the dataset's broad taxonomic coverage and its rich semantic and visual annotation density, making it highly suitable for complex fine-grained visual recognition and reasoning tasks.

*The authors are with the Key Laboratory of AI Safety of CAS, Institute of Computing Technology, Chinese Academy of Sciences (CAS), Beijing 100190, China. E-mail: {ziwei.yao, qiyang.wan}@vipl.ict.ac.cn, {wangruiping, xlchen}@ict.ac.cn*
*\* Corresponding author.*

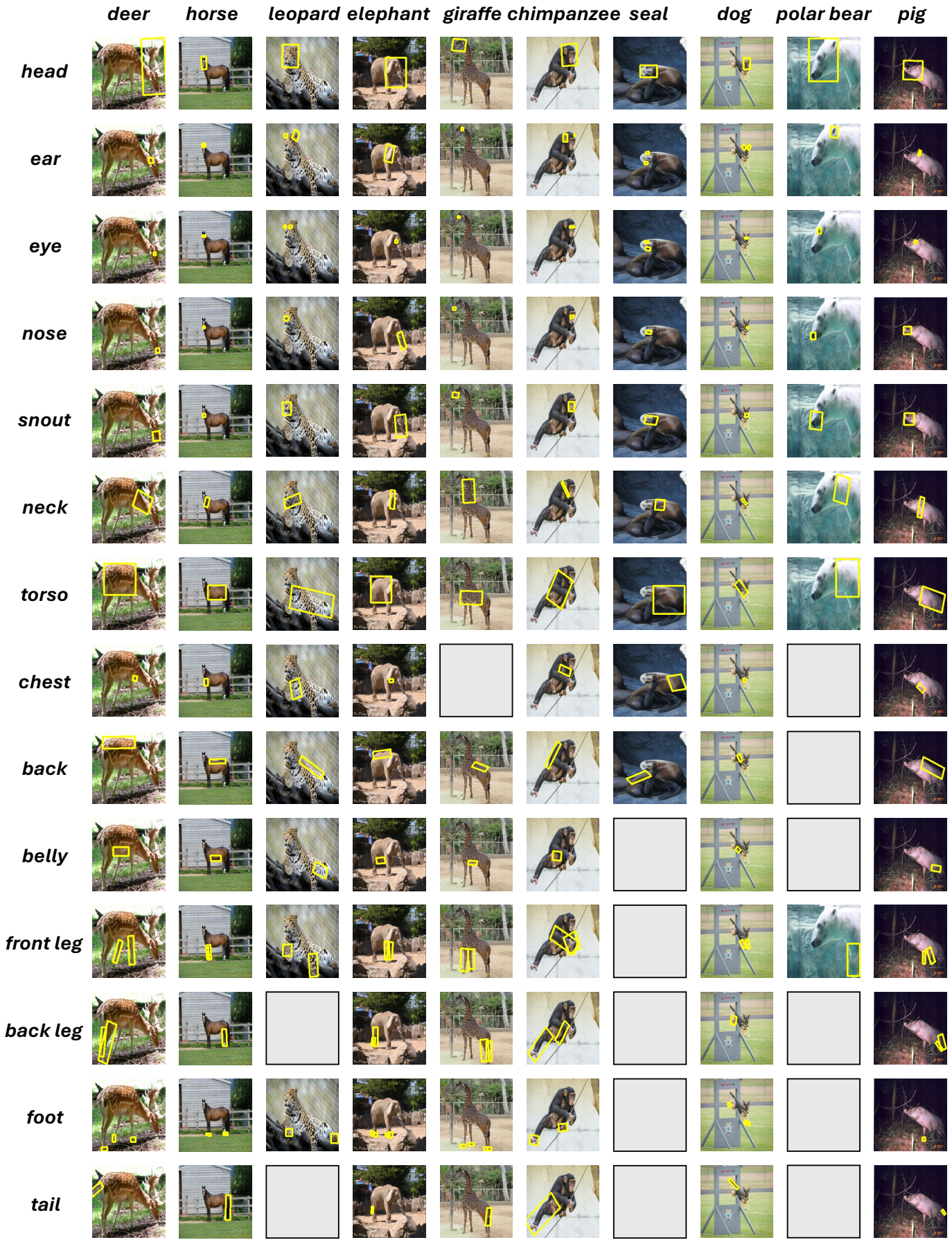

Fig. 1. Visual examples of fine-grained part annotations of categories in AnimalKB.

TABLE 1: Category and part details of AnimalKB.

| major group | sub group | animal category | parts | #triplets | #annot. images | #avg. bboxes |
|---|---|---|---|---|---|---|
| Amphibians | Amphibians | amphibian | - | 102 | - | - |
| Amphibians | Amphibians | african bullfrog | head, eye, mouth, body, leg, foot | 78 | 207 | 5 |
| Amphibians | Amphibians | common platanna | head, eye, mouth, body, leg, foot | 38 | 141 | 5.2 |
| Amphibians | Amphibians | frog | head, eye, mouth, body, leg, foot | 76 | 300 | 7.4 |
| Amphibians | Amphibians | poison frog | head, eye, mouth, body, leg, foot | 35 | 198 | 6.5 |
| Amphibians | Amphibians | salamander | head, eye, mouth, body, leg, foot | 50 | 300 | 5.5 |
| Amphibians | Amphibians | toad | head, eye, mouth, body, leg, foot | 68 | 300 | 5.3 |
| Animals | Animals | animal | - | 105 | - | - |
| Arthropods | Arachnids | arachnid | - | 67 | - | - |
| Arthropods | Arachnids | crab | covering, leg, claw | 72 | 295 | 4 |
| Arthropods | Arachnids | spider | head, body, leg | 58 | 300 | 4.3 |
| Arthropods | Arachnids | tarantula | head, body, leg | 80 | 300 | 4.3 |
| Arthropods | Arachnids | tick and mite | head, body, leg | 45 | 300 | 3.9 |
| Arthropods | Arachnids | scorpion | head, body, leg, tail, claw | 49 | 300 | 5.7 |
| Arthropods | Arthropods | arthropod | - | 101 | - | - |
| Arthropods | Arthropods | centipede and millipede | head, body, leg | 54 | 300 | 3 |
| Arthropods | Crustaceans | crustacean | - | 90 | - | - |
| Arthropods | Crustaceans | krill | head, body, tail, leg, claw | 44 | 245 | 3.3 |
| Arthropods | Crustaceans | lobster | head, body, tail, leg, claw | 47 | 286 | 5.4 |
| Arthropods | Crustaceans | shrimp | head, body, tail, leg, claw | 63 | 286 | 3.1 |
| Arthropods | Insects | insect | - | 117 | - | - |
| Arthropods | Insects | flea | antenna, leg, head, thorax, abdomen | 44 | 272 | 3.8 |
| Arthropods | Insects | walkingstick | antenna, leg, head, thorax, abdomen | 32 | 300 | 3.4 |
| Arthropods | Insects | ant | antenna, leg, head, thorax, abdomen, wing | 58 | 300 | 5.7 |
| Arthropods | Insects | bee | antenna, leg, head, thorax, abdomen, wing | 64 | 300 | 5.5 |
| Arthropods | Insects | beetle | antenna, leg, head, thorax, abdomen, wing | 55 | 300 | 5.6 |
| Arthropods | Insects | butterfly and moth | antenna, leg, head, thorax, abdomen, wing | 79 | 300 | 4.1 |
| Arthropods | Insects | cockroach | antenna, leg, head, thorax, abdomen, wing | 45 | 300 | 5.8 |
| Arthropods | Insects | cricket | antenna, leg, head, thorax, abdomen, wing | 40 | 300 | 7.3 |
| Arthropods | Insects | dragonfly | antenna, leg, head, thorax, abdomen, wing | 44 | 300 | 6.3 |
| Arthropods | Insects | dung beetle | antenna, leg, head, thorax, abdomen, wing | 54 | 109 | 4.5 |
| Arthropods | Insects | firefly | antenna, leg, head, thorax, abdomen, wing | 38 | 264 | 4 |
| Arthropods | Insects | fly | antenna, leg, head, thorax, abdomen, wing | 43 | 300 | 6 |
| Arthropods | Insects | grasshopper | antenna, leg, head, thorax, abdomen, wing | 51 | 300 | 7.2 |

TABLE 1: Category and part details of AnimalKB – Continued

| major group | sub group | animal category | parts | #triplets | #annot. images | #avg. bboxes |
|---|---|---|---|---|---|---|
| Arthropods | Insects | ladybug | antenna, leg, head, thorax, abdomen, wing | 33 | 300 | 4.6 |
| Arthropods | Insects | locust | antenna, leg, head, thorax, abdomen, wing | 32 | 300 | 6 |
| Arthropods | Insects | louse | antenna, leg, head, thorax, abdomen, wing | 48 | 282 | 4.1 |
| Arthropods | Insects | mantis | antenna, leg, head, thorax, abdomen, wing | 29 | 300 | 6.4 |
| Arthropods | Insects | mosquito | antenna, leg, head, thorax, abdomen, wing | 46 | 237 | 5.8 |
| Arthropods | Insects | termite | antenna, leg, head, thorax, abdomen, wing | 38 | 15 | 3.5 |
| Arthropods | Insects | wasp | antenna, leg, head, thorax, abdomen, wing | 43 | 300 | 5.2 |
| Arthropods | Insects | caterpillar | body | 129 | 300 | 1.1 |
| Birds | Birds | bird | - | 92 | - | - |
| Birds | Birds | crow | head, eye, beak, body, wing, foot, tail | 44 | 300 | 9.2 |
| Birds | Birds | magpie | head, eye, beak, body, wing, foot, tail | 47 | 300 | 9.6 |
| Birds | Birds | pied crow | head, eye, beak, body, wing, foot, tail | 48 | 178 | 6.9 |
| Birds | Birds | raven | head, eye, beak, body, wing, foot, tail | 47 | 300 | 8.7 |
| Birds | Birds | warbler | head, eye, beak, body, wing, foot, tail | 56 | 300 | 7.1 |
| Birds | Birds of Prey | bird of prey | - | 38 | - | - |
| Birds | Birds of Prey | african fish eagle | head, eye, beak, body, wing, foot, tail | 47 | 162 | 7.7 |
| Birds | Birds of Prey | bald eagle | head, eye, beak, body, wing, foot, tail | 36 | 300 | 8.2 |
| Birds | Birds of Prey | buzzard | head, eye, beak, body, wing, foot, tail | 47 | 211 | 7.5 |
| Birds | Birds of Prey | eagle | head, eye, beak, body, wing, foot, tail | 62 | 300 | 5.8 |
| Birds | Birds of Prey | falcon | head, eye, beak, body, wing, foot, tail | 46 | 300 | 8.2 |
| Birds | Birds of Prey | hawk | head, eye, beak, body, wing, foot, tail | 57 | 300 | 6.7 |
| Birds | Birds of Prey | lappet-faced vulture | head, eye, beak, body, wing, foot, tail | 53 | 75 | 9.3 |
| Birds | Birds of Prey | martial eagle | head, eye, beak, body, wing, foot, tail | 39 | 186 | 7.5 |
| Birds | Birds of Prey | osprey | head, eye, beak, body, wing, foot, tail | 60 | 300 | 6.8 |
| Birds | Birds of Prey | owl | head, eye, beak, body, wing, foot, tail | 52 | 300 | 9.5 |
| Birds | Birds of Prey | peregrine falcon | head, eye, beak, body, wing, foot, tail | 41 | 172 | 8.4 |
| Birds | Birds of Prey | secretary bird | head, eye, beak, body, wing, foot, tail | 48 | 125 | 9.2 |
| Birds | Birds of Prey | verreaux's eagle | head, eye, beak, body, wing, foot, tail | 42 | 74 | 5.1 |
| Birds | Birds of Prey | vulture | head, eye, beak, body, wing, foot, tail | 51 | 300 | 10.1 |

TABLE 1: Category and part details of AnimalKB – Continued

| major group | sub group | animal category | parts | #triplets | #annot. images | #avg. bboxes |
|---|---|---|---|---|---|---|
| Birds | Flightless Birds | flightless birds | - | 43 | - | - |
| Birds | Flightless Birds | african penguin | head, eye, beak, body, wing, foot, tail | 42 | 179 | 9.8 |
| Birds | Flightless Birds | emu | head, eye, beak, body, wing, foot, tail | 39 | 230 | 9.1 |
| Birds | Flightless Birds | kiwi | head, eye, beak, body, wing, foot, tail | 47 | 71 | 6.4 |
| Birds | Flightless Birds | ostrich | head, eye, beak, body, wing, foot, tail | 54 | 300 | 10.2 |
| Birds | Flightless Birds | penguin | head, eye, beak, body, wing, foot, tail | 54 | 300 | 13.8 |
| Birds | Flightless Birds | rhea | head, eye, beak, body, wing, foot, tail | 46 | 182 | 9.3 |
| Birds | Other Birds | albatross | head, eye, beak, body, wing, foot, tail | 52 | 300 | 9.1 |
| Birds | Other Birds | cowbird | head, eye, beak, body, wing, foot, tail | 32 | 300 | 8 |
| Birds | Other Birds | cuckoo | head, eye, beak, body, wing, foot, tail | 34 | 300 | 7.8 |
| Birds | Other Birds | hummingbird | head, eye, beak, body, wing, foot, tail | 43 | 300 | 8.8 |
| Birds | Other Birds | kingfisher | head, eye, beak, body, wing, foot, tail | 34 | 300 | 6.6 |
| Birds | Other Birds | kookaburra | head, eye, beak, body, wing, foot, tail | 43 | 178 | 6.9 |
| Birds | Other Birds | kori bustard | head, eye, beak, body, wing, foot, tail | 58 | 141 | 8.2 |
| Birds | Other Birds | parrot family | head, eye, beak, body, wing, foot, tail | 61 | 300 | 9.9 |
| Birds | Other Birds | peacock | head, eye, beak, body, wing, foot, tail | 52 | 300 | 10.4 |
| Birds | Other Birds | pigeon and dove | head, eye, beak, body, wing, foot, tail | 71 | 300 | 10.6 |
| Birds | Other Birds | quetzal | head, eye, beak, body, wing, foot, tail | 80 | 194 | 6.6 |
| Birds | Other Birds | roadrunner | head, eye, beak, body, wing, foot, tail | 46 | 300 | 7.2 |
| Birds | Other Birds | southern ground hornbill | head, eye, beak, body, wing, foot, tail | 44 | 164 | 8.7 |
| Birds | Other Birds | toucan | head, eye, beak, body, wing, foot, tail | 99 | 300 | 8.2 |
| Birds | Other Birds | whippoorwill | head, eye, beak, body, wing, foot, tail | 39 | 67 | 5.1 |
| Birds | Other Birds | woodpecker | head, eye, beak, body, wing, foot, tail | 45 | 300 | 7.3 |
| Birds | Poultry | poultry | - | 13 | - | - |
| Birds | Poultry | chicken | head, eye, beak, body, wing, foot, tail | 50 | 299 | 11.5 |
| Birds | Poultry | guinea fowl | head, eye, beak, body, wing, foot, tail | 62 | 168 | 8.4 |
| Birds | Poultry | turkey | head, eye, beak, body, wing, foot, tail | 55 | 300 | 10.9 |
| Birds | Shorebirds | gull | head, eye, beak, body, wing, foot, tail | 59 | 300 | 9.3 |
| Birds | Shorebirds | loon | head, eye, beak, body, wing, foot, tail | 53 | 300 | 5.8 |

TABLE 1: Category and part details of AnimalKB – Continued

| major group | sub group | animal category | parts | #triplets | #annot. images | #avg. bboxes |
|---|---|---|---|---|---|---|
| Birds | Shorebirds | pelican | head, eye, beak, body, wing, foot, tail | 43 | 300 | 11.1 |
| Birds | Shorebirds | sandpiper | head, eye, beak, body, wing, foot, tail | 84 | 300 | 9.7 |
| Birds | Shorebirds | tern | head, eye, beak, body, wing, foot, tail | 51 | 300 | 10.8 |
| Birds | Songbirds | songbird | - | 28 | - | - |
| Birds | Songbirds | blackbird | head, eye, beak, body, wing, foot, tail | 72 | 300 | 7.4 |
| Birds | Songbirds | bluebird | head, eye, beak, body, wing, foot, tail | 51 | 242 | 8.6 |
| Birds | Songbirds | canary | head, eye, beak, body, wing, foot, tail | 31 | 216 | 9.6 |
| Birds | Songbirds | cardinal | head, eye, beak, body, wing, foot, tail | 51 | 179 | 7.8 |
| Birds | Songbirds | finch | head, eye, beak, body, wing, foot, tail | 61 | 300 | 9.3 |
| Birds | Songbirds | mockingbird | head, eye, beak, body, wing, foot, tail | 45 | 300 | 7.6 |
| Birds | Songbirds | nightingale | head, eye, beak, body, wing, foot, tail | 41 | 80 | 8.4 |
| Birds | Songbirds | robin | head, eye, beak, body, wing, foot, tail | 76 | 300 | 9.1 |
| Birds | Songbirds | sparrow | head, eye, beak, body, wing, foot, tail | 66 | 300 | 7.5 |
| Birds | Wading Birds | blue crane | head, eye, beak, body, wing, foot, tail | 47 | 190 | 8.3 |
| Birds | Wading Birds | crane | head, eye, beak, body, wing, foot, tail | 72 | 296 | 9.5 |
| Birds | Wading Birds | flamingo | head, eye, beak, body, wing, foot, tail | 57 | 300 | 11.2 |
| Birds | Wading Birds | hadada ibis | head, eye, beak, body, wing, foot, tail | 51 | 171 | 7.5 |
| Birds | Wading Birds | heron | head, eye, beak, body, wing, foot, tail | 54 | 300 | 8.9 |
| Birds | Wading Birds | ibis | head, eye, beak, body, wing, foot, tail | 54 | 300 | 9.2 |
| Birds | Wading Birds | stork | head, eye, beak, body, wing, foot, tail | 57 | 300 | 9.3 |
| Birds | Waterfowl | duck | head, eye, beak, body, wing, foot, tail | 87 | 300 | 9.4 |
| Birds | Waterfowl | goose | head, eye, beak, body, wing, foot, tail | 55 | 300 | 11.1 |
| Birds | Waterfowl | goose egyptian | head, eye, beak, body, wing, foot, tail | 60 | 175 | 11.5 |
| Birds | Waterfowl | swan | head, eye, beak, body, wing, foot, tail | 62 | 300 | 8.8 |
| Extinct Animals | Extinct Animals | Allosaurus | - | 56 | - | - |
| Extinct Animals | Extinct Animals | Ankylosaurus | - | 47 | - | - |
| Extinct Animals | Extinct Animals | Apatosaurus | - | 17 | - | - |
| Extinct Animals | Extinct Animals | Archaeopteryx | - | 24 | - | - |
| Extinct Animals | Extinct Animals | Brachiosaurus | - | 47 | - | - |
| Extinct Animals | Extinct Animals | Ceratosaurus | - | 81 | - | - |
| Extinct Animals | Extinct Animals | Compsognathus | - | 40 | - | - |
| Extinct Animals | Extinct Animals | Deinonychus | - | 63 | - | - |

TABLE 1: Category and part details of AnimalKB – Continued

| major group | sub group | animal category | parts | #triplets | #annot. images | #avg. bboxes |
|---|---|---|---|---|---|---|
| Extinct Animals | Extinct Animals | dinosaur | - | 143 | - | - |
| Extinct Animals | Extinct Animals | Diplodocus | - | 69 | - | - |
| Extinct Animals | Extinct Animals | Elasmosaurus | - | 27 | - | - |
| Extinct Animals | Extinct Animals | Eoraptor | - | 30 | - | - |
| Extinct Animals | Extinct Animals | Euoplocephalus | - | 39 | - | - |
| Extinct Animals | Extinct Animals | Extinct Anima | - | 24 | - | - |
| Extinct Animals | Extinct Animals | fossil | - | 22 | - | - |
| Extinct Animals | Extinct Animals | Heterodontosaurus | - | 64 | - | - |
| Extinct Animals | Extinct Animals | Hypsilophodon | - | 37 | - | - |
| Extinct Animals | Extinct Animals | Ichthyosaurus | - | 33 | - | - |
| Extinct Animals | Extinct Animals | Iguanodon | - | 45 | - | - |
| Extinct Animals | Extinct Animals | Kronosaurus | - | 34 | - | - |
| Extinct Animals | Extinct Animals | Lambeosaurus | - | 22 | - | - |
| Extinct Animals | Extinct Animals | Maiasaura | - | 46 | - | - |
| Extinct Animals | Extinct Animals | Megalosaurus | - | 59 | - | - |
| Extinct Animals | Extinct Animals | Mrs. Ples | - | 21 | - | - |
| Extinct Animals | Extinct Animals | Oviraptor | - | 25 | - | - |
| Extinct Animals | Extinct Animals | Pachycephalosaurus | - | 69 | - | - |
| Extinct Animals | Extinct Animals | Plateosaurus | - | 47 | - | - |
| Extinct Animals | Extinct Animals | plesiosaur | - | 46 | - | - |
| Extinct Animals | Extinct Animals | prehistoric life | - | 117 | - | - |
| Extinct Animals | Extinct Animals | Protoceratops | - | 48 | - | - |
| Extinct Animals | Extinct Animals | Psittacosaurus | - | 32 | - | - |
| Extinct Animals | Extinct Animals | Pteranodon | - | 47 | - | - |
| Extinct Animals | Extinct Animals | pterodactyl | - | 28 | - | - |
| Extinct Animals | Extinct Animals | Stegosaurus | - | 53 | - | - |
| Extinct Animals | Extinct Animals | Triceratops | - | 50 | - | - |
| Extinct Animals | Extinct Animals | Tyrannosaurus Rex | - | 48 | - | - |
| Extinct Animals | Extinct Animals | Velociraptor | - | 39 | - | - |
| Fish | Fish | fish | - | 76 | - | - |
| Fish | Fish | anchovy | head, body, tail, fin, gill, eye, mouth | 40 | 48 | 6.3 |
| Fish | Fish | angelfish | head, body, tail, fin, gill, eye, mouth | 46 | 98 | 4.4 |
| Fish | Fish | barracuda | head, body, tail, fin, gill, eye, mouth | 36 | 212 | 4.7 |
| Fish | Fish | bass | head, body, tail, fin, gill, eye, mouth | 44 | 290 | 6.6 |
| Fish | Fish | butterfly fish | head, body, tail, fin, gill, eye, mouth | 40 | 116 | 4.2 |
| Fish | Fish | carp | head, body, tail, fin, gill, eye, mouth | 31 | 281 | 7.3 |
| Fish | Fish | catfish | head, body, tail, fin, gill, eye, mouth | 34 | 163 | 7.9 |
| Fish | Fish | clown fish | head, body, tail, fin, gill, eye, mouth | 61 | 122 | 4.4 |
| Fish | Fish | cod | head, body, tail, fin, gill, eye, mouth | 32 | 27 | 9.4 |
| Fish | Fish | coelacanth | head, body, tail, fin, gill, eye, mouth | 24 | 239 | 7.2 |
| Fish | Fish | eel | head, body, tail, fin, gill, eye, mouth | 64 | 300 | 5.8 |
| Fish | Fish | flying fish | head, body, tail, fin, gill, eye, mouth | 26 | 264 | 5 |
| Fish | Fish | galjoen | head, body, tail, fin, gill, eye, mouth | 24 | 89 | 4.8 |

TABLE 1: Category and part details of AnimalKB – Continued

| major group | sub group | animal category | parts | #triplets | #annot. images | #avg. bboxes |
|---|---|---|---|---|---|---|
| Fish | Fish | goldfish | head, body, tail, fin, gill, eye, mouth | 40 | 300 | 6.9 |
| Fish | Fish | great white shark | head, body, tail, fin, gill, eye, mouth | 58 | 300 | 8.2 |
| Fish | Fish | guppy | head, body, tail, fin, gill, eye, mouth | 55 | 120 | 4.8 |
| Fish | Fish | herring | head, body, tail, fin, gill, eye, mouth | 34 | 12 | 7.8 |
| Fish | Fish | lamprey | head, body, tail, fin, gill, eye, mouth | 40 | 240 | 4.1 |
| Fish | Fish | parrot fish | head, body, tail, fin, gill, eye, mouth | 61 | 246 | 4.2 |
| Fish | Fish | piranha | head, body, tail, fin, gill, eye, mouth | 36 | 92 | 7.4 |
| Fish | Fish | puffer | head, body, tail, fin, gill, eye, mouth | 52 | 300 | 7.1 |
| Fish | Fish | ray | head, body, tail, fin, gill, eye, mouth | 49 | 300 | 6.4 |
| Fish | Fish | sailfish | head, body, tail, fin, gill, eye, mouth | 42 | 262 | 6.6 |
| Fish | Fish | salmon | head, body, tail, fin, gill, eye, mouth | 50 | 300 | 9.3 |
| Fish | Fish | sea horse | head, body, tail, fin, gill, eye, mouth | 45 | 224 | 4.2 |
| Fish | Fish | shark | head, body, tail, fin, gill, eye, mouth | 93 | 300 | 7.4 |
| Fish | Fish | snoek | head, body, tail, fin, gill, eye, mouth | 28 | 300 | 8.2 |
| Fish | Fish | sturgeon | head, body, tail, fin, gill, eye, mouth | 88 | 300 | 9.8 |
| Fish | Fish | swordfish | head, body, tail, fin, gill, eye, mouth | 38 | 202 | 7.5 |
| Fish | Fish | trout | head, body, tail, fin, gill, eye, mouth | 45 | 300 | 6.4 |
| Fish | Fish | tuna | head, body, tail, fin, gill, eye, mouth | 40 | 279 | 6.8 |
| Fish | Fish | whale shark | head, body, tail, fin, gill, eye, mouth | 46 | 163 | 6.5 |
| Mammals | Felines | cat | - | 85 | - | - |
| Mammals | Felines | saber-toothed cat | - | 73 | - | - |
| Mammals | Felines | bobcat | head, ear, eye, snout, torso, leg, foot, tail | 97 | 574 | 13 |
| Mammals | Felines | cheetah | head, ear, eye, snout, torso, leg, foot, tail | 61 | 300 | 13.9 |
| Mammals | Felines | jaguar | head, ear, eye, snout, torso, leg, foot, tail | 43 | 300 | 11.4 |
| Mammals | Felines | leopard | head, ear, eye, snout, torso, leg, foot, tail | 95 | 671 | 12.1 |
| Mammals | Felines | lynx | head, ear, eye, snout, torso, leg, foot, tail | 64 | 300 | 14.4 |
| Mammals | Felines | ocelot | head, ear, eye, snout, torso, leg, foot, tail | 60 | 68 | 11.1 |
| Mammals | Felines | puma | head, ear, eye, snout, torso, leg, foot, tail | 36 | 300 | 13.1 |
| Mammals | Felines | tiger | head, ear, eye, snout, torso, leg, foot, tail | 102 | 811 | 13.6 |

TABLE 1: Category and part details of AnimalKB – Continued

| major group | sub group | animal category | parts | #triplets | #annot. images | #avg. bboxes |
|---|---|---|---|---|---|---|
| Mammals | Felines | lion | head, ear, eye, snout, torso, leg, foot, tail, mane | 92 | 786 | 13.5 |
| Mammals | Canines | dog | - | 107 | - | - |
| Mammals | Canines | african hunting dog | head, ear, eye, snout, torso, leg, foot, tail | 51 | 300 | 14.7 |
| Mammals | Canines | arctic fox | head, ear, eye, snout, torso, leg, foot, tail | 48 | 300 | 12.6 |
| Mammals | Canines | basset hound | head, ear, eye, snout, torso, leg, foot, tail | 28 | 300 | 13.8 |
| Mammals | Canines | beagle | head, ear, eye, snout, torso, leg, foot, tail | 26 | 300 | 13.1 |
| Mammals | Canines | bernese mountain dog | head, ear, eye, snout, torso, leg, foot, tail | 47 | 300 | 14.5 |
| Mammals | Canines | border collie | head, ear, eye, snout, torso, leg, foot, tail | 25 | 300 | 13.8 |
| Mammals | Canines | boxer | head, ear, eye, snout, torso, leg, foot, tail | 46 | 300 | 14.4 |
| Mammals | Canines | bulldog | head, ear, eye, snout, torso, leg, foot, tail | 52 | 300 | 13.8 |
| Mammals | Canines | collie | head, ear, eye, snout, torso, leg, foot, tail | 77 | 851 | 12.8 |
| Mammals | Canines | coyote | head, ear, eye, snout, torso, leg, foot, tail | 40 | 300 | 13.7 |
| Mammals | Canines | dachshund | head, ear, eye, snout, torso, leg, foot, tail | 31 | 89 | 13.5 |
| Mammals | Canines | dalmatian | head, ear, eye, snout, torso, leg, foot, tail | 64 | 451 | 14.6 |
| Mammals | Canines | dingo | head, ear, eye, snout, torso, leg, foot, tail | 48 | 300 | 15 |
| Mammals | Canines | doberman pinscher | head, ear, eye, snout, torso, leg, foot, tail | 36 | 300 | 14.5 |
| Mammals | Canines | fox | head, ear, eye, snout, torso, leg, foot, tail | 78 | 612 | 15.2 |
| Mammals | Canines | foxhound | head, ear, eye, snout, torso, leg, foot, tail | 27 | 300 | 14.7 |
| Mammals | Canines | german shepherd | head, ear, eye, snout, torso, leg, foot, tail | 71 | 789 | 13.8 |
| Mammals | Canines | golden retriever | head, ear, eye, snout, torso, leg, foot, tail | 28 | 300 | 12.7 |
| Mammals | Canines | greyhound | head, ear, eye, snout, torso, leg, foot, tail | 33 | 300 | 14.4 |
| Mammals | Canines | irish setter | head, ear, eye, snout, torso, leg, foot, tail | 25 | 300 | 13.8 |
| Mammals | Canines | jackal | head, ear, eye, snout, torso, leg, foot, tail | 52 | 64 | 13.6 |
| Mammals | Canines | labrador retriever | head, ear, eye, snout, torso, leg, foot, tail | 30 | 300 | 13.6 |
| Mammals | Canines | maltese | head, ear, eye, snout, torso, leg, foot, tail | 24 | 300 | 12.4 |
| Mammals | Canines | old english sheepdog | head, ear, eye, snout, torso, leg, foot, tail | 33 | 300 | 12.8 |
| Mammals | Canines | pit bull terrier | head, ear, eye, snout, torso, leg, foot, tail | 27 | 300 | 15.2 |
| Mammals | Canines | pomeranian | head, ear, eye, snout, torso, leg, foot, tail | 25 | 300 | 12.5 |

TABLE 1: Category and part details of AnimalKB – Continued

| major group | sub group | animal category | parts | #triplets | #annot. images | #avg. bboxes |
|---|---|---|---|---|---|---|
| Mammals | Canines | poodle | head, ear, eye, snout, torso, leg, foot, tail | 35 | 290 | 13.3 |
| Mammals | Canines | saint bernard | head, ear, eye, snout, torso, leg, foot, tail | 29 | 300 | 14.1 |
| Mammals | Canines | sheepdog | head, ear, eye, snout, torso, leg, foot, tail | 8 | 300 | 13.9 |
| Mammals | Canines | shetland sheepdog | head, ear, eye, snout, torso, leg, foot, tail | 26 | 300 | 15.3 |
| Mammals | Canines | spaniel | head, ear, eye, snout, torso, leg, foot, tail | 63 | 300 | 13.4 |
| Mammals | Canines | terrier | head, ear, eye, snout, torso, leg, foot, tail | 28 | 300 | 13.4 |
| Mammals | Canines | welsh corgi | head, ear, eye, snout, torso, leg, foot, tail | 47 | 300 | 15.2 |
| Mammals | Canines | wolf | head, ear, eye, snout, torso, leg, foot, tail | 89 | 472 | 13.9 |
| Mammals | Egg-Laying Mammals | echidna | head, ear, eye, snout, torso, leg, foot, tail | 55 | 300 | 7.3 |
| Mammals | Egg-Laying Mammals | platypus | head, ear, eye, snout, torso, leg, foot, tail | 50 | 300 | 7.3 |
| Mammals | Hoofed Mammals | woolly rhinoceros | - | 43 | - | - |
| Mammals | Hoofed Mammals | donkey | head, ear, eye, snout, torso, leg, foot, tail | 33 | 238 | 15.6 |
| Mammals | Hoofed Mammals | peccary | head, ear, eye, snout, torso, leg, foot, tail | 56 | 179 | 10.5 |
| Mammals | Hoofed Mammals | tapir | head, ear, eye, snout, torso, leg, foot, tail | 50 | 29 | 12.9 |
| Mammals | Hoofed Mammals | zebra | head, ear, eye, snout, torso, leg, foot, tail | 66 | 630 | 16 |
| Mammals | Hoofed Mammals | deer | head, ear, eye, snout, torso, leg, foot, tail, antler | 105 | 914 | 14.6 |
| Mammals | Hoofed Mammals | moose | head, ear, eye, snout, torso, leg, foot, tail, antler | 88 | 502 | 13.8 |
| Mammals | Hoofed Mammals | reindeer | head, ear, eye, snout, torso, leg, foot, tail, antler | 47 | 250 | 11.9 |
| Mammals | Hoofed Mammals | wapiti | head, ear, eye, snout, torso, leg, foot, tail, antler, neck | 38 | 196 | 14.9 |
| Mammals | Hoofed Mammals | antelope | head, ear, eye, snout, torso, leg, foot, tail, horn | 78 | 590 | 17 |
| Mammals | Hoofed Mammals | bison | head, ear, eye, snout, torso, leg, foot, tail, horn | 63 | 300 | 14.6 |
| Mammals | Hoofed Mammals | buffalo | head, ear, eye, snout, torso, leg, foot, tail, horn | 86 | 424 | 12.9 |
| Mammals | Hoofed Mammals | cattle | head, ear, eye, snout, torso, leg, foot, tail, horn | 97 | 300 | 18.2 |
| Mammals | Hoofed Mammals | gemsbok | head, ear, eye, snout, torso, leg, foot, tail, horn | 50 | 37 | 16.8 |
| Mammals | Hoofed Mammals | goat | head, ear, eye, snout, torso, leg, foot, tail, horn | 55 | 300 | 15.6 |
| Mammals | Hoofed Mammals | impala | head, ear, eye, snout, torso, leg, foot, tail, horn | 65 | 300 | 17.4 |
| Mammals | Hoofed Mammals | kudu | head, ear, eye, snout, torso, leg, foot, tail, horn | 72 | 52 | 14.9 |
| Mammals | Hoofed Mammals | mountain goat | head, ear, eye, snout, torso, leg, foot, tail, horn | 60 | 81 | 14.6 |

TABLE 1: Category and part details of AnimalKB – Continued

| major group | sub group | animal category | parts | #triplets | #annot. images | #avg. bboxes |
|---|---|---|---|---|---|---|
| Mammals | Hoofed Mammals | musk-ox | head, ear, eye, snout, torso, leg, foot, tail, horn | 53 | 119 | 9.8 |
| Mammals | Hoofed Mammals | rhinoceros | head, ear, eye, snout, torso, leg, foot, tail, horn | 88 | 519 | 15.5 |
| Mammals | Hoofed Mammals | sheep | head, ear, eye, snout, torso, leg, foot, tail, horn | 83 | 435 | 14.6 |
| Mammals | Hoofed Mammals | springbok | head, ear, eye, snout, torso, leg, foot, tail, horn | 49 | 85 | 15.9 |
| Mammals | Hoofed Mammals | wildebeest | head, ear, eye, snout, torso, leg, foot, tail, horn | 58 | 49 | 16.2 |
| Mammals | Hoofed Mammals | yak | head, ear, eye, snout, torso, leg, foot, tail, horn | 60 | 88 | 14.2 |
| Mammals | Hoofed Mammals | eland | head, ear, eye, snout, torso, leg, foot, tail, horn, neck | 54 | 186 | 14.3 |
| Mammals | Hoofed Mammals | giraffe | head, ear, eye, snout, torso, leg, foot, tail, horn, neck | 73 | 854 | 12.7 |
| Mammals | Hoofed Mammals | horse | head, ear, eye, snout, torso, leg, foot, tail, mane | 105 | 839 | 13.9 |
| Mammals | Hoofed Mammals | alpaca | head, ear, eye, snout, torso, leg, foot, tail, neck | 44 | 300 | 15.8 |
| Mammals | Hoofed Mammals | camel | head, ear, eye, snout, torso, leg, foot, tail, neck | 74 | 300 | 20.4 |
| Mammals | Hoofed Mammals | guanaco | head, ear, eye, snout, torso, leg, foot, tail, neck | 52 | 95 | 15.3 |
| Mammals | Hoofed Mammals | llama | head, ear, eye, snout, torso, leg, foot, tail, neck | 42 | 300 | 17.1 |
| Mammals | Hoofed Mammals | okapi | head, ear, eye, snout, torso, leg, foot, tail, neck | 42 | 11 | 17.7 |
| Mammals | Hoofed Mammals | vicuna | head, ear, eye, snout, torso, leg, foot, tail, neck | 47 | 152 | 15.1 |
| Mammals | Hoofed Mammals | hippopotamus | head, ear, eye, snout, torso, leg, foot, tail, tusk | 62 | 436 | 10.9 |
| Mammals | Hoofed Mammals | pig | head, ear, eye, snout, torso, leg, foot, tail, tusk | 108 | 414 | 13.1 |
| Mammals | Hoofed Mammals | warthog | head, ear, eye, snout, torso, leg, foot, tail, tusk | 70 | 300 | 17.6 |
| Mammals | Marsupials | kangaroo | - | 41 | - | - |
| Mammals | Marsupials | marsupial | - | 29 | - | - |
| Mammals | Marsupials | bilby | head, ear, eye, snout, torso, leg, foot, tail | 56 | 99 | 11.6 |
| Mammals | Marsupials | koala | head, ear, eye, snout, torso, leg, foot, tail | 43 | 300 | 12.7 |
| Mammals | Marsupials | opossum | head, ear, eye, snout, torso, leg, foot, tail | 43 | 300 | 8.7 |
| Mammals | Marsupials | tasmanian devil | head, ear, eye, snout, torso, leg, foot, tail | 40 | 189 | 11.2 |
| Mammals | Marsupials | wallaby | head, ear, eye, snout, torso, leg, foot, tail | 56 | 300 | 15.6 |
| Mammals | Marsupials | wombat | head, ear, eye, snout, torso, leg, foot, tail | 65 | 300 | 12.9 |
| Mammals | Mammals | mammal | - | 139 | - | - |
| Mammals | Mammals | quagga | - | 26 | - | - |
| Mammals | Other Mammals | Mastodon and Mammoth | - | 31 | - | - |
| Mammals | Other Mammals | aardvark | head, ear, eye, snout, torso, leg, foot, tail | 37 | 213 | 9.4 |

TABLE 1: Category and part details of AnimalKB – Continued

| major group | sub group | animal category | parts | #triplets | #annot. images | #avg. bboxes |
|---|---|---|---|---|---|---|
| Mammals | Other Mammals | anteater | head, ear, eye, snout, torso, leg, foot, tail | 75 | 113 | 8.9 |
| Mammals | Other Mammals | armadillo | head, ear, eye, snout, torso, leg, foot, tail | 47 | 300 | 8.4 |
| Mammals | Other Mammals | badger | head, ear, eye, snout, torso, leg, foot, tail | 62 | 300 | 12 |
| Mammals | Other Mammals | bat | head, ear, eye, snout, torso, leg, foot, tail | 90 | 291 | 11 |
| Mammals | Other Mammals | bear | head, ear, eye, snout, torso, leg, foot, tail | 103 | 300 | 12.9 |
| Mammals | Other Mammals | hedgehog | head, ear, eye, snout, torso, leg, foot, tail | 49 | 300 | 9.5 |
| Mammals | Other Mammals | honey badger | head, ear, eye, snout, torso, leg, foot, tail | 55 | 117 | 8.3 |
| Mammals | Other Mammals | hyena | head, ear, eye, snout, torso, leg, foot, tail | 50 | 300 | 14 |
| Mammals | Other Mammals | hyrax | head, ear, eye, snout, torso, leg, foot, tail | 74 | 174 | 9.1 |
| Mammals | Other Mammals | meerkat | head, ear, eye, snout, torso, leg, foot, tail | 46 | 300 | 12.4 |
| Mammals | Other Mammals | mink | head, ear, eye, snout, torso, leg, foot, tail | 48 | 300 | 11.6 |
| Mammals | Other Mammals | mole | head, ear, eye, snout, torso, leg, foot, tail | 76 | 99 | 10.1 |
| Mammals | Other Mammals | mongoose | head, ear, eye, snout, torso, leg, foot, tail | 45 | 300 | 13.4 |
| Mammals | Other Mammals | panda | head, ear, eye, snout, torso, leg, foot, tail | 103 | 748 | 11.3 |
| Mammals | Other Mammals | pangolin | head, ear, eye, snout, torso, leg, foot, tail | 57 | 120 | 5.8 |
| Mammals | Other Mammals | polar bear | head, ear, eye, snout, torso, leg, foot, tail | 89 | 767 | 12.5 |
| Mammals | Other Mammals | rabbit and hare | head, ear, eye, snout, torso, leg, foot, tail | 82 | 300 | 11 |
| Mammals | Other Mammals | raccoon | head, ear, eye, snout, torso, leg, foot, tail | 92 | 397 | 12.5 |
| Mammals | Other Mammals | riverine rabbit | head, ear, eye, snout, torso, leg, foot, tail | 34 | 75 | 10.8 |
| Mammals | Other Mammals | shrew | head, ear, eye, snout, torso, leg, foot, tail | 41 | 235 | 5.9 |
| Mammals | Other Mammals | skunk | head, ear, eye, snout, torso, leg, foot, tail | 74 | 150 | 11.1 |
| Mammals | Other Mammals | sloth | head, ear, eye, snout, torso, leg, foot, tail | 53 | 300 | 9.2 |
| Mammals | Other Mammals | wolverine | head, ear, eye, snout, torso, leg, foot, tail | 42 | 193 | 7.9 |
| Mammals | Other Mammals | otter | head, ear, eye, snout, torso, leg, foot, tail, neck | 80 | 489 | 10.7 |
| Mammals | Other Mammals | weasel | head, ear, eye, snout, torso, leg, foot, tail, neck | 76 | 236 | 12.3 |
| Mammals | Other Mammals | elephant | head, ear, eye, snout, torso, leg, foot, tail, tusk | 99 | 654 | 15 |
| Mammals | Primates | primate | - | 52 | - | - |
| Mammals | Primates | ape | head, ear, eye, snout, torso, leg, foot, arm, hand | 100 | 300 | 12.9 |

TABLE 1: Category and part details of AnimalKB – Continued

| major group | sub group | animal category | parts | #triplets | #annot. images | #avg. bboxes |
|---|---|---|---|---|---|---|
| Mammals | Primates | bonobo | head, ear, eye, snout, torso, leg, foot, arm, hand | 39 | 135 | 14.9 |
| Mammals | Primates | chimpanzee | head, ear, eye, snout, torso, leg, foot, arm, hand | 94 | 490 | 12.2 |
| Mammals | Primates | gibbon | head, ear, eye, snout, torso, leg, foot, arm, hand | 53 | 300 | 12.9 |
| Mammals | Primates | gorilla | head, ear, eye, snout, torso, leg, foot, arm, hand | 116 | 739 | 12.2 |
| Mammals | Primates | loris | head, ear, eye, snout, torso, leg, foot, arm, hand | 60 | 250 | 7.8 |
| Mammals | Primates | orangutan | head, ear, eye, snout, torso, leg, foot, arm, hand | 52 | 300 | 12.6 |
| Mammals | Primates | baboon | head, ear, eye, snout, torso, leg, foot, tail, arm, hand | 87 | 300 | 16.5 |
| Mammals | Primates | chacma baboon | head, ear, eye, snout, torso, leg, foot, tail, arm, hand | 43 | 171 | 11.5 |
| Mammals | Primates | howler monkey | head, ear, eye, snout, torso, leg, foot, tail, arm, hand | 46 | 300 | 12.3 |
| Mammals | Primates | lemur | head, ear, eye, snout, torso, leg, foot, tail, arm, hand | 61 | 300 | 14.6 |
| Mammals | Primates | mandrill | head, ear, eye, snout, torso, leg, foot, tail, arm, hand | 60 | 120 | 12.2 |
| Mammals | Primates | marmoset | head, ear, eye, snout, torso, leg, foot, tail, arm, hand | 53 | 300 | 13.1 |
| Mammals | Primates | monkey | head, ear, eye, snout, torso, leg, foot, tail, arm, hand | 126 | 296 | 13.6 |
| Mammals | Primates | spider monkey | head, ear, eye, snout, torso, leg, foot, tail, arm, hand | 39 | 300 | 14.7 |
| Mammals | Primates | tarsier | head, ear, eye, snout, torso, leg, foot, tail, arm, hand | 54 | 172 | 12.1 |
| Mammals | Rodents | rodent | - | 40 | - | - |
| Mammals | Rodents | beaver | head, ear, eye, snout, torso, leg, foot, tail | 97 | 154 | 9.9 |
| Mammals | Rodents | chipmunk | head, ear, eye, snout, torso, leg, foot, tail | 64 | 125 | 11 |
| Mammals | Rodents | gerbil | head, ear, eye, snout, torso, leg, foot, tail | 49 | 124 | 10.3 |
| Mammals | Rodents | groundhog | head, ear, eye, snout, torso, leg, foot, tail | 28 | 62 | 10.7 |
| Mammals | Rodents | guinea pig | head, ear, eye, snout, torso, leg, foot, tail | 56 | 300 | 12 |
| Mammals | Rodents | hamster | head, ear, eye, snout, torso, leg, foot, tail | 90 | 706 | 10.5 |
| Mammals | Rodents | lemming | head, ear, eye, snout, torso, leg, foot, tail | 33 | 163 | 8.2 |
| Mammals | Rodents | marmot | head, ear, eye, snout, torso, leg, foot, tail | 59 | 300 | 11.1 |
| Mammals | Rodents | mouse | head, ear, eye, snout, torso, leg, foot, tail | 62 | 158 | 12.3 |
| Mammals | Rodents | muskrat | head, ear, eye, snout, torso, leg, foot, tail | 40 | 33 | 8.8 |
| Mammals | Rodents | porcupine | head, ear, eye, snout, torso, leg, foot, tail | 45 | 300 | 9.2 |
| Mammals | Rodents | prairie dog | head, ear, eye, snout, torso, leg, foot, tail | 38 | 45 | 10.9 |

TABLE 1: Category and part details of AnimalKB – Continued

| major group | sub group | animal category | parts | #triplets | #annot. images | #avg. bboxes |
|---|---|---|---|---|---|---|
| Mammals | Rodents | rat | head, ear, eye, snout, torso, leg, foot, tail | 66 | 234 | 11.8 |
| Mammals | Rodents | squirrel | head, ear, eye, snout, torso, leg, foot, tail | 86 | 1187 | 15.6 |
| Mammals | Rodents | vole | head, ear, eye, snout, torso, leg, foot, tail | 47 | 48 | 6.9 |
| Mammals | Sea Mammals | blue whale | head, body, tail, fin, blowhole, eye, mouth | 71 | 25 | 3.6 |
| Mammals | Sea Mammals | dolphin | head, body, tail, fin, blowhole, eye, mouth | 73 | 511 | 7.5 |
| Mammals | Sea Mammals | killer whale | head, body, tail, fin, blowhole, eye, mouth | 93 | 158 | 5.9 |
| Mammals | Sea Mammals | porpoise | head, body, tail, fin, blowhole, eye, mouth | 43 | 250 | 5.3 |
| Mammals | Sea Mammals | southern right whale | head, body, tail, fin, blowhole, eye, mouth | 60 | 182 | 4.5 |
| Mammals | Sea Mammals | whale | head, body, tail, fin, blowhole, eye, mouth | 100 | 300 | 6.1 |
| Mammals | Sea Mammals | manatee | head, ear, eye, snout, torso, flipper, tail | 25 | 114 | 5.3 |
| Mammals | Sea Mammals | sea lion | head, ear, eye, snout, torso, flipper, tail | 74 | 300 | 10.1 |
| Mammals | Sea Mammals | seal | head, ear, eye, snout, torso, flipper, tail | 79 | 569 | 12.2 |
| Mammals | Sea Mammals | walrus | head, ear, eye, snout, torso, flipper, tail, tusk | 84 | 135 | 13.3 |
| Mollusks | Mollusks | mollusk | - | 92 | - | - |
| Mollusks | Mollusks | octopus | body, eye, leg | 49 | 300 | 4.4 |
| Mollusks | Mollusks | giant squid | body, eye, leg, fin | 36 | 70 | 5 |
| Mollusks | Mollusks | squid | body, eye, leg, fin | 45 | 128 | 4.3 |
| Mollusks | Mollusks | bivalve | body, shell | 47 | 266 | 3.7 |
| Mollusks | Mollusks | clam | body, shell | 74 | 261 | 4.6 |
| Mollusks | Mollusks | oyster | body, shell | 89 | 271 | 4.3 |
| Mollusks | Mollusks | scallop | body, shell | 78 | 236 | 2.9 |
| Mollusks | Mollusks | snail and slug | body, shell | 63 | 300 | 1.6 |
| Other Animals | Other Animals | amoeba | - | 28 | - | - |
| Other Animals | Other Animals | Great Barrier Reef | | 33 | - | - |
| Other Animals | Other Animals | hydra | - | 22 | - | - |
| Other Animals | Other Animals | protozoan | - | 38 | - | - |
| Other Animals | Other Animals | coral | body | 57 | 300 | 1.7 |
| Other Animals | Other Animals | sea anemone | body | 51 | 300 | 2 |
| Other Animals | Other Animals | sea cucumber | body | 48 | 300 | 1.8 |
| Other Animals | Other Animals | sea star | body | 47 | 300 | 1.6 |
| Other Animals | Other Animals | sea urchin | body | 44 | 145 | 2.8 |
| Other Animals | Other Animals | worm | body | 76 | 300 | 1.5 |
| Other Animals | Other Animals | sponge | body, hole | 30 | 201 | 4.5 |
| Other Animals | Other Animals | jellyfish | body, leg | 48 | 300 | 3.7 |
| Reptiles | Reptiles | reptile | - | 122 | - | - |
| Reptiles | Reptiles | adder | head, eye, mouth, body | 72 | 250 | 2.6 |
| Reptiles | Reptiles | anaconda | head, eye, mouth, body | 56 | 217 | 3.1 |
| Reptiles | Reptiles | boa constrictor | head, eye, mouth, body | 40 | 300 | 3.7 |
| Reptiles | Reptiles | boomslang | head, eye, mouth, body | 34 | 194 | 3 |
| Reptiles | Reptiles | cape cobra | head, eye, mouth, body | 29 | 137 | 2.8 |
| Reptiles | Reptiles | cobra | head, eye, mouth, body | 48 | 300 | 3.3 |
| Reptiles | Reptiles | copperhead | head, eye, mouth, body | 69 | 300 | 1.8 |

TABLE 1: Category and part details of AnimalKB – Continued

| major group | sub group | animal category | parts | #triplets | #annot. images | #avg. bboxes |
|---|---|---|---|---|---|---|
| Reptiles | Reptiles | coral snake | head, eye, mouth, body | 39 | 279 | 2.3 |
| Reptiles | Reptiles | garter snake | head, eye, mouth, body | 41 | 300 | 2.8 |
| Reptiles | Reptiles | mamba | head, eye, mouth, body | 42 | 300 | 3.1 |
| Reptiles | Reptiles | moccasin | head, eye, mouth, body | 60 | 227 | 2.4 |
| Reptiles | Reptiles | puff adder | head, eye, mouth, body | 32 | 300 | 2.5 |
| Reptiles | Reptiles | python | head, eye, mouth, body | 37 | 300 | 2.9 |
| Reptiles | Reptiles | racer | head, eye, mouth, body | 54 | 300 | 2.4 |
| Reptiles | Reptiles | rat snake | head, eye, mouth, body | 83 | 300 | 2.3 |
| Reptiles | Reptiles | rattlesnake | head, eye, mouth, body | 45 | 300 | 2.9 |
| Reptiles | Reptiles | sea snake | head, eye, mouth, body | 39 | 300 | 2.5 |
| Reptiles | Reptiles | snake | head, eye, mouth, body | 115 | 300 | 2.4 |
| Reptiles | Reptiles | viper | head, eye, mouth, body | 47 | 300 | 2.8 |
| Reptiles | Reptiles | water snake | head, eye, mouth, body | 58 | 300 | 3 |
| Reptiles | Reptiles | alligator | head, eye, mouth, body, leg, foot, tail | 86 | 300 | 8.1 |
| Reptiles | Reptiles | chameleon | head, eye, mouth, body, leg, foot, tail | 36 | 300 | 8.6 |
| Reptiles | Reptiles | crocodile | head, eye, mouth, body, leg, foot, tail | 65 | 300 | 7.1 |
| Reptiles | Reptiles | gecko | head, eye, mouth, body, leg, foot, tail | 40 | 300 | 6.4 |
| Reptiles | Reptiles | geometric tortoise | head, eye, mouth, body, leg, foot, tail | 34 | 66 | 4.8 |
| Reptiles | Reptiles | gila monster | head, eye, mouth, body, leg, foot, tail | 51 | 300 | 6.1 |
| Reptiles | Reptiles | iguana | head, eye, mouth, body, leg, foot, tail | 50 | 300 | 10.6 |
| Reptiles | Reptiles | komodo dragon | head, eye, mouth, body, leg, foot, tail | 57 | 300 | 10.4 |
| Reptiles | Reptiles | leatherback turtle | head, eye, mouth, body, leg, foot, tail | 48 | 300 | 7.2 |
| Reptiles | Reptiles | lizard | head, eye, mouth, body, leg, foot, tail | 120 | 300 | 8.2 |
| Reptiles | Reptiles | nile crocodile | head, eye, mouth, body, leg, foot, tail | 50 | 300 | 8.3 |
| Reptiles | Reptiles | sea turtle | head, eye, mouth, body, leg, foot, tail | 79 | 300 | 7.3 |
| Reptiles | Reptiles | turtle | head, eye, mouth, body, leg, foot, tail | 58 | 300 | 7.1 |

## 2 KNOWLEDGE VISUALIZATION OF ANIMALKB

Here we provide some instances of knowledge visualization from AnimalKB, as shown in the Fig. 2. On the left is the visualized text knowledge graph corresponding to animal categories, centered around the node representing the animal itself, where all entities are interlinked. The graphs are constructed by connecting the triplets from AnimalKB. On the right are the corresponding images and local regions. In AnimalKB, the text-based knowledge and the image-based knowledge are aligned, providing a comprehensive understanding of object categories.

## 3 DETAILS OF DOWNSTREAM TASKS

In this section, we provide details on the application of AnimalKB in downstream tasks. This includes a more detailed discussion on zero-shot recognition and additional experimental results in fine-grained VQA.

### 3.1 Discussion on Zero-shot Recognition

In the field of computer vision, a series of proxy tasks – such as image recognition, object detection, and semantic segmentation – are commonly employed to evaluate the capabilities of models. Mainstream methodologies generally exhibit robust performance when trained on large-scale labeled datasets. However, their effectiveness declines substantially when confronted with novel categories for which no annotated samples are available during training. The process of adapting a recognition model from known to unknown categories is formally defined as zero-shot learning (ZSL). In the context of object recognition, ZSL specifically aims to develop models capable of accurately classifying instances from unseen categories, despite the absence of labeled training data pertaining to these categories during the training phase. To enable this capability, the incorporation of auxiliary external knowledge is typically required, thereby enhancing cross-domain generalization and knowledge transfer within the model.

Classic zero-shot recognition methods are motivated by the challenge of identifying objects or concepts that were not encountered during the model's training phase. Traditional supervised learning approaches require labeled examples for every category to be recognized, which significantly limits their scalability and applicability to real-world scenarios where new categories continually emerge. To address this limitation, ZSL facilitates knowledge transfer from seen (training) categories to unseen (test) categories by exploiting auxiliary information, such as semantic embeddings or attribute descriptions. These methods often rely on a shared semantic space, in which both seen and unseen categories are represented, thus enabling the model to generalize to novel classes by understanding their relationships to known ones. For example, if a model is trained on animals such as "dog" and "cat," it can infer the characteristics of an unseen animal like "zebra" by leveraging semantic similarities or attribute descriptions (e.g., "stripes" or "four legs").

Nevertheless, in recent years, the advent of vision-language models such as CLIP [1] has revolutionized zero-shot recognition. This setting is often referred to as "pseudo" zero-shot because, although the model has not been explicitly trained on the unseen categories, these categories may have been indirectly encountered during CLIP's pre-training phase. For example, if CLIP was pre-trained on a dataset containing images of "zebras," the model has already seen examples of this category, even if "zebra" was not included in the downstream task's training data. This circumstance raises questions about the "true" zero-shot nature of such methods, as the model benefits from prior exposure to novel categories during pre-training. However, CLIP-based approaches have become the mainstream in zero-shot recognition due to their impressive generalization capabilities and ease of use, setting a new benchmark for the field. Consequently, we adopt this setting as our experimental setup in the main paper.

This approach is particularly advantageous as it capitalizes on the rich semantic knowledge embedded within the pre-trained model, thereby enabling effective generalization to the common categories based solely on their textual descriptions. While open-vocabulary models demonstrate strong capabilities in generating embeddings from class names, their performance is frequently constrained when class labels are coarsely defined or lack sufficient semantic granularity. This limitation is particularly evident in datasets with implicit hierarchical structures, where coarse labels fail to adequately capture the nuanced relationships between categories.

In the main paper, we transform the annotated triplets for each animal category into natural language phrases, which serve as fine-grained descriptions for CLIP to compute similarity with images. These descriptions enrich the semantic information of the categories and enhance the model's ability to discern subtle differences between classes by introducing more specific contextual relationships. By converting structured knowledge into natural language descriptions, we preserve the semantic hierarchy of the original data and enable a deeper exploration of categorical information, thereby providing more robust support for zero-shot recognition in complex scenarios. This refinement improves the alignment between images and text, resulting in enhanced zero-shot classification performance.

TABLE 2
Performance of closed-source MLLMs on VQA with AnimalKB.

| Model | Visual | Non-visual | All |
|---|---|---|---|
| **without KB** | | | |
| gpt-4o (1120) | 78.95 | 76.81 | 77.88 |
| gpt-4o (0806) | 79.36 | 77.13 | 78.25 |
| gpt-4o-mini | 69.78 | 66.86 | 68.32 |
| gemini-2.0-flash | **84.95** | **81.96** | **83.46** |
| gemini-1.5-pro | 78.68 | 78.68 | 79.71 |
| claude-3.5-sonnet | 81.05 | 77.82 | 79.44 |
| claude-3-haiku | 63.82 | 59.14 | 61.48 |
| **with KB** | | | |
| gpt-4o (1120) | 98.26 | 97.87 | 98.06 |
| gpt-4o (0806) | **98.68** | **97.79** | **98.24** |
| gpt-4o-mini | 94.04 | 95.07 | 94.56 |
| gemini-2.0-flash | 98.16 | 97.65 | 97.90 |
| gemini-1.5-pro | 97.01 | 96.54 | 96.78 |
| claude-3.5-sonnet | 97.43 | 96.72 | 97.07 |
| claude-3-haiku | 92.06 | 92.21 | 92.13 |

leopard



hawk



Fig. 2. Illustration of more examples from the AnimalKB. The left part shows fully connected knowledge graphs centered on animal nodes, whereas the right part displays image region annotations on both animal and part level.

### 3.2 Additional Results of Fine-grained VQA

In the main paper, due to space limitations, we present only a subset of the evaluation results for both closed-source and open-source multi-modal models on the Fine-grained VQA task. Here, we provide the complete set of results, as shown in Tab. 2 and Tab. 3. With the table with full results, the conclusions from the main paper are upheld: the integration of knowledge bases significantly enhances model performance, and is closely linked to the scale of the models' parameters.

Additionally, due to limited context length of Instruct-BLIP [2], it is not feasible to input information for both the correct and distracting categories simultaneously, as is done with other models. Consequently, we only provide information for the correct category, resulting in a simpler task, named "VQA with Ground-truth KB". To ensure a fair comparison, we evaluate the performance of various models under this setting as well. The utilization of a ground-truth knowledge base significantly enhances the performance of open-source models, as demonstrated in Tab. 4. Models such as `InternVL2.5-8B` and `Qwen2-VL-7B` achieve accuracies of 89.38% and 89.08%, respectively, when employing a ground-truth KB. This finding highlights the critical importance of both the existence and quality of knowledge bases, as well as the retrieval accuracy of knowledge-retrieval

models in practical applications. Models that effectively leverage high-quality KBs are likely to achieve superior performance.

## 4 PROMPTS

### 4.1 Prompts for Construction

In automated textual knowledge construction, we utilize the LLM to extract knowledge triples from a given text under certain constraints. For each knowledge triple, we analyze the animal or part entities it contains and determine whether the knowledge is visual or non-visual. The prompt is shown in Fig. 3.

### 4.2 Prompts in Applications

In the fine-grained VQA task, we develop a reanswering procedure transitioning from a "without KB" to a "with KB" setting. The primary distinction for the evaluated MLLMs is that, under the "without KB" condition, the model's prompt comprises only the input image, task description, question, and answer options; in contrast, the "with KB" prompt additionally incorporates four sets of knowledge entries utilized for RAG. The relevant prompts during these two phases are depicted in Fig. 4 and Fig. 5 respectively. Moreover, we evaluate LLM-based methods on the task of

TABLE 3
Performance of open-source MLLMs on VQA with AnimalKB.

| Model | Size | Visual | Non-visual | All |
|---|---|---|---|---|
| **without KB** | | | | |
| DeepSeek-VL2 | 27.5 B | 73.75 | **_75.69_** | **_74.72_** |
| DeepSeek-VL2-Small | 16.1 B | 66.13 | 68.70 | 67.41 |
| DeepSeek-VL2-Tiny | 3.4 B | 64.02 | 62.06 | 63.04 |
| Qwen2.5-VL-7B | 8.3 B | 74.44 | 71.25 | 72.84 |
| Qwen2-VL-7B | 8.3 B | 74.56 | 72.84 | 73.70 |
| InternVL2.5-8B | 8.1 B | 71.27 | 69.09 | 70.18 |
| InternVL2.5-26B | 25.5 B | **_76.52_** | 72.55 | 74.53 |
| InternVL2-8B | 8.0 B | 71.08 | 68.97 | 70.02 |
| InternVL2-26B | 25.5 B | 72.61 | 69.03 | 70.82 |
| LLaVA-v1.6-Mistral-7B | 7.6 B | 65.54 | 60.10 | 62.82 |
| LLaVA-v1.6-Vicuna-7B | 7.1 B | 65.76 | 60.91 | 63.33 |
| LLaVA-v1.6-Vicuna-13B | 13.4 B | 64.75 | 62.70 | 63.73 |
| LLaVA-Next-Llama3-8B | 8.4 B | 65.47 | 63.75 | 64.61 |
| LLaVA-OV-Qwen2-7B | 8.0 B | 71.94 | 65.27 | 68.60 |
| **with KB** | | | | |
| DeepSeek-VL2 | 27.5 B | 84.66 | 85.10 | 84.88 |
| DeepSeek-VL2-Small | 16.1 B | 85.39 | 84.73 | 85.06 |
| DeepSeek-VL2-Tiny | 3.4 B | 71.96 | 72.84 | 72.40 |
| Qwen2.5-VL-7B | 8.3 B | 87.06 | 85.42 | 86.24 |
| Qwen2-VL-7B | 8.3 B | 87.67 | 86.91 | 87.29 |
| InternVL2.5-8B | 8.1 B | 87.23 | 86.47 | 86.85 |
| InternVL2.5-26B | 25.5 B | **_87.99_** | **_88.04_** | **_88.01_** |
| InternVL2-8B | 8.0 B | 86.27 | 85.20 | 85.74 |
| InternVL2-26B | 25.5 B | 87.16 | 85.29 | 86.23 |
| LLaVA-v1.6-Mistral-7B | 7.6 B | 82.11 | 81.81 | 81.96 |
| LLaVA-v1.6-Vicuna-7B | 7.1 B | 80.71 | 80.86 | 80.78 |
| LLaVA-v1.6-Vicuna-13B | 13.4 B | 82.13 | 81.74 | 81.94 |
| LLaVA-Next-Llama3-8B | 8.4 B | 83.41 | 84.19 | 83.80 |
| LLaVA-OV-Qwen2-7B | 8.0 B | 83.28 | 81.81 | 82.55 |

TABLE 4
Performance of open-source MLLMs on VQA with Ground-truth KB.

| Model | Size | Visual | Non-visual | All |
|---|---|---|---|---|
| **without KB** | | | | |
| Qwen2.5-VL-7B | 8.3 B | 74.44 | 71.25 | 72.84 |
| Qwen2-VL-7B | 8.3 B | **_74.56_** | **_72.84_** | **_73.70_** |
| InternVL2.5-8B | 8.1 B | 71.27 | 69.09 | 70.18 |
| InternVL2-8B | 8.0 B | 71.08 | 68.97 | 70.02 |
| LLaVA-Next-Llama3-8B | 8.4 B | 65.47 | 63.75 | 64.61 |
| InstructBLIP-Vicuna-7B | 7.9 B | 47.87 | 47.03 | 47.45 |
| InstructBLIP-Vicuna-13B | 14.2 B | 53.97 | 54.73 | 54.35 |
| InstructBLIP-Flan-T5-XL | 4.0 B | 62.16 | 59.63 | 60.89 |
| InstructBLIP-Flan-T5-XXL | 12.3 B | 66.27 | 62.03 | 64.15 |
| **with Ground Truth KB** | | | | |
| Qwen2.5-VL-7B | 8.3 B | 88.21 | 87.87 | 88.04 |
| Qwen2-VL-7B | 8.3 B | 89.26 | 88.90 | 89.08 |
| InternVL2.5-8B | 8.1 B | **_89.73_** | **_89.02_** | **_89.38_** |
| InternVL2-8B | 8.0 B | 88.60 | 87.99 | 88.30 |
| LLaVA-Next-Llama3-8b | 8.4 B | 87.87 | 88.92 | 88.39 |
| InstructBLIP-Vicuna-7B | 7.9 B | 67.08 | 69.07 | 68.08 |
| InstructBLIP-Vicuna-13B | 14.2 B | 76.27 | 78.65 | 77.46 |
| InstructBLIP-Flan-T5-XL | 4.0 B | 84.93 | 82.65 | 83.79 |
| InstructBLIP-Flan-T5-XXL | 12.3 B | 87.67 | 87.18 | 87.43 |

Knowledge Graph Completion (KGC). The prompt used in the LLM-based link prediction is provided in Fig. 6.

# 5 SYSTEMS

## 5.1 Manual Knowledge Extraction System

We build a complete set of web-based system for manual knowledge extraction, containing textual knowledge extraction, seed image annotation, and quality inspection, respec-

---

Define 2 Knowledge Types:
1. visual knowledge: physical features of the animal or its parts, such as appearance, part numbers, etc.
2. non-visual knowledge: knowledge NOT related to animal appearance, such as category, behavior, habitat, food, etc.
Given a paragraph, for each sentence, please find the knowledge types included in the sentence and extract knowledge triplets based on knowledge types. Output a list of triplets. For visual knowledge, provide [Entity1, Visual Relation, Entity2], Visual Relation come from the list of visual relations. For non-visual knowledge, provide [Entity1, Non-visual Relation, Entity2], Visual Relation come from the list of non-visual relations.

Notice that the number and appearance of animal part should be considered as visual knowledge!
Type number and time, should be considered as non-visual knowledge!
The pronouns should be replaced with what they refer to!
The parallel items in one knowledge should be split into different triplets!
If the knowledge involves animal parts (including feathers, hair, etc.), you MUST break down the parts and their attributes as much as possible. The attribute of parts, like number, color, size and others, should be considered as Entity2 of that part, instead of adjunct word before the part. For each part, provide triplet [animal, Have, part] and triplets should begin with the part. For example, triplets of "eagles have two brown legs" should be [eagle, Have, leg], [leg, Num, 2], [leg, Color, brown].
For each sentence, output the sentence, knowledge type it contains and a list of knowledge triplets. Only output results!

[Paragraph]
The African fish eagle is a species, or type, of sea eagle.
Sea eagles are large eagles that feed on fish.

[Output]
1. The African fish eagle is a species, or type, of sea eagle.
Knowledge types: non-visual knowledge
Triplets: [African fish eagle, BelongTo, sea eagle]
2. Sea eagles are large eagles that feed on fish.
Knowledge types: visual knowledge, non-visual knowledge
Triplets: [sea eagle, Size, large], [sea eagle, Eat, fish]

[Paragraph]
{Text to extract}
[Output]

Fig. 3. Prompt of textual knowledge extraction.

tively. The screenshots of their user interfaces are shown in Fig. 7.

## 5.2 Visual Knowledge Base Exploration System

For better presentation of the comprehensive content of the visual knowledge base and ease of use, we build a

You are an expert in zoology. Your job is to recognize the animal in the image and answer multiple-choice questions w.r.t. it.

You will be provided with an image of an animal species and several multiple-choice questions. Each question will include a paragraph and four options labeled "A," "B," "C," and "D." Please recognize the species of the animal, answer the questions and output the label of the correct option for each question.

The questions are given in the following form:
- question_text, [A. option_A, B. option_B, C. option_C, D. option_D]
Your answer should be in the JSON format.

Now, the image and corresponding questions are listed as follows:
{image}
{questions}

Fig. 4. Prompt of **answering** without AnimalKB.

knowledge graph exploration system based on React and the force graph, as shown in the Fig. 8. The left side of the page features category selection, the middle section is the KG display area, and the right side provides detailed information, including animal encyclopedia articles, triples, and related images with annotations. When nodes and edges in the KG are clicked, the right side will display information related to that instance. Image annotations are displayed in a carousel format, automatically rotating.

You are an expert in zoology. Your job is to recognize the animal in the image and answer multiple-choice questions w.r.t. it.

You will be provided with an image of an animal species and several multiple-choice questions. Each question will include a paragraph and four options labeled "A," "B," "C," and "D." Please recognize the species of the animal, answer the questions and output the label of the correct option for each question.

Here is some concept knowledge about certain animal species (which may be relevant or irrelevant to the questions) presented in the form of triples for your reference:
- {knowledge_of_correct_concept}
- {knowledge_of_interfering_concept_1}
- {knowledge_of_interfering_concept_2}
- {knowledge_of_interfering_concept_3}

The questions are given in the following form:
- question_text, [A. option_A, B. option_B, C. option_C, D. option_D]
Your answer should be in the JSON format.

Now, the image and corresponding questions are listed as follows:
{image}
{questions}

Fig. 5. Prompt of **re-answering** with AnimalKB.

You are an expert in the field of animals. Your job is to help construct and complete an animal knowledge graph.
You will be provided with an incomplete triplet from an animal knowledge graph, along with a list of candidate entities. Your task is to predict the most likely head or tail entity in the triplet to complete missing information.

Input format:
- An incomplete triplet *[Entity1, Relation, MASK]* or *[MASK, Relation, Entity2]*: Predict the missing head entity or tail entity ('MASK').
- A list of candidate entities.

Only output the sorted list of candidate entities as the json format, ranked from most likely to least likely.

Now, rank the missing entity for the incomplete triplet.
{Incomplete triplet}
{Candidate entities}

Fig. 6. Prompt of LLM-based link prediction

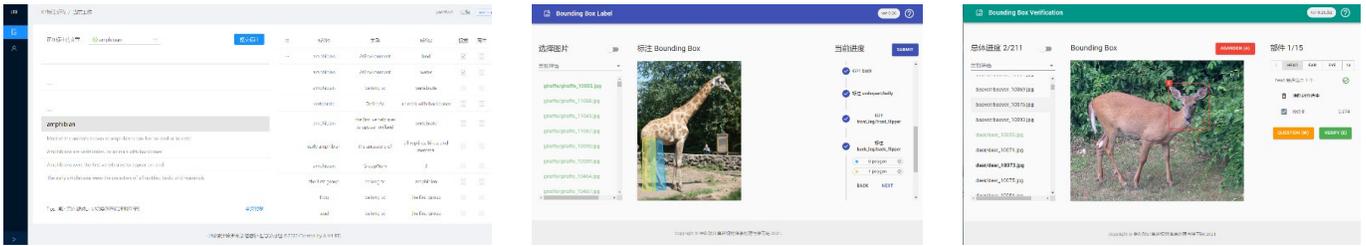

Fig. 7. User interface of the manual knowledge extraction systems, including the text-source knowledge extraction system, seed image annotation system, and quality inspection system.

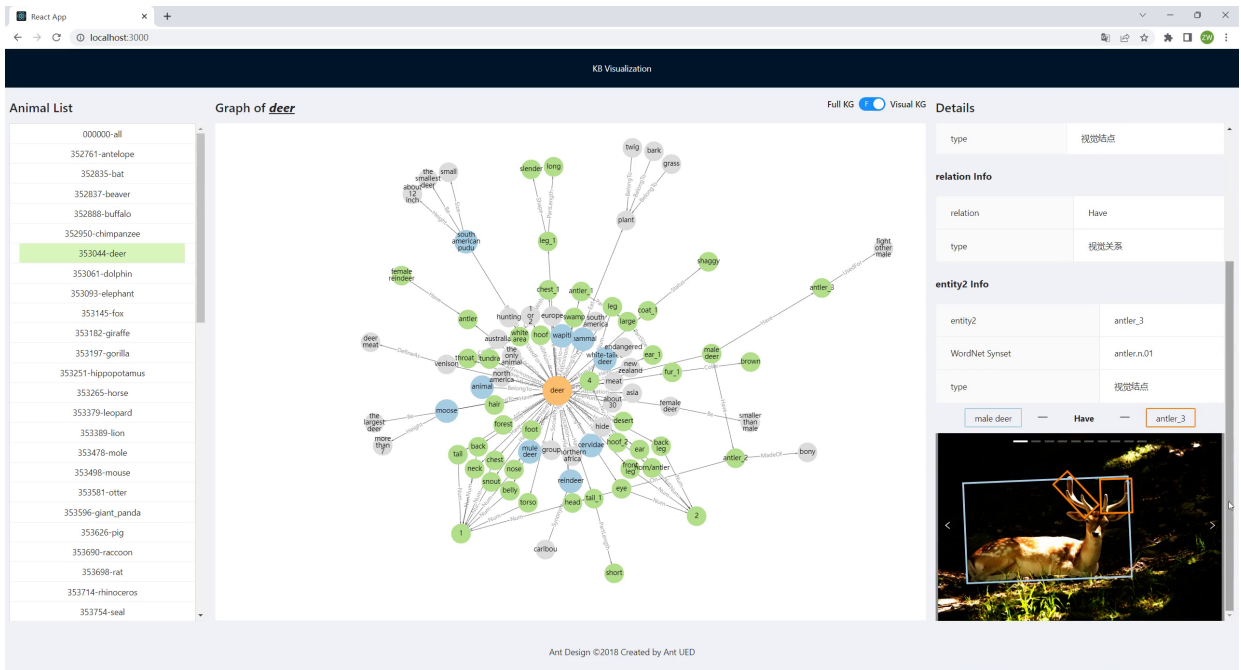

Fig. 8. User interface of the visual knowledge base exploration system, which contains the visualization of the knowledge graph, detailed triple information, and corresponding images with region annotations.



\end{document}